\documentclass[10pt,twocolumn,letterpaper]{article}

\usepackage{config/iccv2023/iccv}

\usepackage{graphicx}

\usepackage{amsmath}
\usepackage{amssymb}
\usepackage{times}
\usepackage{epsfig}
\usepackage{xcolor}
\usepackage{datetime2}
\usepackage{bm}
\usepackage{color,soul}          %
\usepackage[nopatch=eqnum]{microtype}      %
\usepackage{dcolumn}        %
\usepackage{url}            %
\usepackage{booktabs}       %
\usepackage{amsfonts}       %
\usepackage{nicefrac}       %
\usepackage{enumitem}
\usepackage{multirow}
\usepackage{currfile}
\usepackage{float}
\usepackage{algorithm}
\usepackage[noend]{algpseudocode}
\usepackage{amsthm}
\usepackage{subfiles}
\usepackage{adjustbox}
\usepackage{subcaption}
\usepackage{makecell}
\usepackage[normalem]{ulem}
\usepackage{caption}
\usepackage[flushmargin,para]{footmisc}
\usepackage{pifont}%
\usepackage{adjustbox}
\usepackage[numbers,sort,compress]{natbib}
\usepackage[pagebackref=true,breaklinks=true,colorlinks,bookmarks=false]{hyperref}

\usepackage[accsupp]{axessibility}  %

\usepackage{wrapfig}

\usepackage{hhline}

\usepackage{balance}

\usepackage{colortbl}  %

\usepackage{fontawesome}
\usepackage{listings}

\usepackage[capitalize]{cleveref}
\crefname{section}{Sec.}{Secs.}
\Crefname{section}{Section}{Sections}
\Crefname{table}{Table}{Tables}
\crefname{table}{Tab.}{Tabs.}
\Crefname{equation}{Equation}{Equations}
\crefname{equation}{Eq.}{Eqs.}

\usepackage[normalem]{ulem}

\usepackage{xspace}
\usepackage{currfile}
\xspaceaddexceptions{]\}}

\newcommand{\camready}[1]{\textcolor{black}{{#1}}\xspace}

\newcommand{\nameCOLOR}[1]{\textcolor{black}{#1}} %
\newcommand{\nameMethod}{\nameCOLOR{\mbox{DECO}}\xspace} %
\newcommand{\nameMethodLong}{\nameCOLOR{\textbf{D}ense \textbf{E}stimation of 3D human-scene \textbf{CO}ntact in the wild}}

\newcommand{\nameDataset}{\nameCOLOR{\mbox{DAMON}}\xspace}
\newcommand{\nameDatasetFull}{\nameCOLOR{\textbf{D}ense \textbf{A}nnotation of \threeD hu\textbf{M}an \textbf{O}bject contact in \textbf{N}atural images}\xspace}
\newcommand{\vcoco}{\nameCOLOR{\mbox{V-COCO}}\xspace}
\newcommand{\hake}{\nameCOLOR{\mbox{HAKE}}\xspace}
\newcommand{\hoi}{\mbox{\nameCOLOR{HOI}}\xspace}
\newcommand{\hsi}{\mbox{\nameCOLOR{HSI}}\xspace}

\newcommand{\projectURL}{\url{https://deco.is.tue.mpg.de}}

\newcommand{\outTitleFULL}{DECO: Dense Estimation of 3D Human-Scene Contact In The Wild}

\newcommand{\video}{\nameCOLOR{\textbf{video}}\xspace}
\newcommand{\supmat}{\nameCOLOR{Sup.~Mat.}\xspace}

\newcommand{\bs}[1]{\boldsymbol{{#1}}}

\usepackage{lipsum}

\newcommand{\smpl}{\mbox{SMPL}\xspace}
\newcommand{\smplx}{\mbox{SMPL-X}\xspace}

\newcommand{\smplh}{\mbox{SMPL-H}\xspace}

\newcommand{\hps}{\mbox{HPS}\xspace}
\newcommand{\HPS}{\hps}

\renewcommand{\etc}{etc\xspace}
\renewcommand{\etal}{\mbox{et al.}\xspace}
\renewcommand{\ie}{\mbox{i.e.}\xspace}
\renewcommand{\eg}{\mbox{e.g.}\xspace}
\renewcommand{\wrt}{\mbox{w.r.t.}\xspace}

\newcommand{\pose}{\boldsymbol{\theta}\xspace}
\newcommand{\shape}{\boldsymbol{\beta}\xspace}

\newcommand{\zheading}[1]{\textbf{#1:}}
\newcommand{\qheading}[1]{\noindent\textbf{#1:}}

\newcommand{\twoD}{{2D}\xspace}

\newcommand{\threeD}{\xspace{3D}\xspace}

\newcommand{\mocap}{\mbox{MoCap}\xspace}

\newcommand{\colorRef}[1]{\textcolor{red}{#1}} %
\usepackage[capitalize]{cleveref}
\crefname{figure}{\colorRef{Fig.}}{\colorRef{Figs.}}
\Crefname{figure}{\colorRef{Figure}}{\colorRef{Figures}}
\crefname{section}{\colorRef{Sec.}}{\colorRef{Secs.}}
\Crefname{section}{\colorRef{Section}}{\colorRef{Sections}}
\Crefname{table}{\colorRef{Table}}{\colorRef{Tables}}
\crefname{table}{\colorRef{Tab.}}{\colorRef{Tabs.}}
\Crefname{equation}{\colorRef{Equation}}{\colorRef{Equations}}
\crefname{equation}{\colorRef{Eq.}}{\colorRef{Eqs.}}

\newcommand{\smplmesh}{\mathcal{M}}
\newcommand{\smplmesht}{\mathcal{\bar{M}}}

\newcommand{\meshPartSET}{\mathcal{P}}
\newcommand{\meshPart}{P}
\newcommand{\meshPartk}{\meshPart_k}

\newcommand{\camrot}{\mathbf{R}^c}
\newcommand{\bodytransl}{\mathbf{t}^b}

\newcommand{\fpart}{\boldsymbol{F_p}}
\newcommand{\fscene}{\boldsymbol{F_s}}
\newcommand{\fcontact}{\boldsymbol{F_c}}

\definecolor{lightgray}{gray}{0.97}
\definecolor{lightblue}{rgb}{0.93,0.95,1.0}

\definecolor{GreenColor}{rgb}{0.137,0.573,0.565}
\definecolor{OrangeColor}{rgb}{0.914,0.541,0.0.141}
\definecolor{PurpleColor}{rgb}{0.5,0,0.7}

\newlist{todolist}{itemize}{2}
\setlist[todolist]{label=$\square$}

\newcommand{\cmark}{\ding{51}}%
\newcommand{\xmark}{\ding{55}}%

\def\iccvPaperID{1035} %

\iccvfinalcopy %

\ificcvfinal\fi

\begin{document}

\title{\vspace{-0.5 em}\outTitleFULL\vspace{-0.5 em}}

\author{
Shashank Tripathi$^{1}$\footnotemark[1]~\footnotemark[2] \quad 
Agniv Chatterjee$^{1}$\footnotemark[1] \quad 
Jean-Claude Passy$^1$ \quad 
Hongwei Yi$^1$ \quad \\
Dimitrios Tzionas$^2$ \quad 
Michael J. Black$^1$ \\
{\small
$^1$Max Planck Institute for Intelligent Systems, T{\"u}bingen, Germany \quad
$^2$University of Amsterdam, the Netherlands
}\\
{\tt\small \{stripathi, achatterjee, jpassy, hyi, black\}@tue.mpg.de \quad {d.tzionas@uva.nl}}\\
}

\twocolumn[{%
\renewcommand\twocolumn[1][]{#1}%
\maketitle
\thispagestyle{empty}

\definecolor{ContactBlue}{rgb}{0.25, 0.48, 0.7}
\begin{center}
    \vspace{-0.5 em}
    \centering
    \captionsetup{type=figure}
    \includegraphics[width=\linewidth]{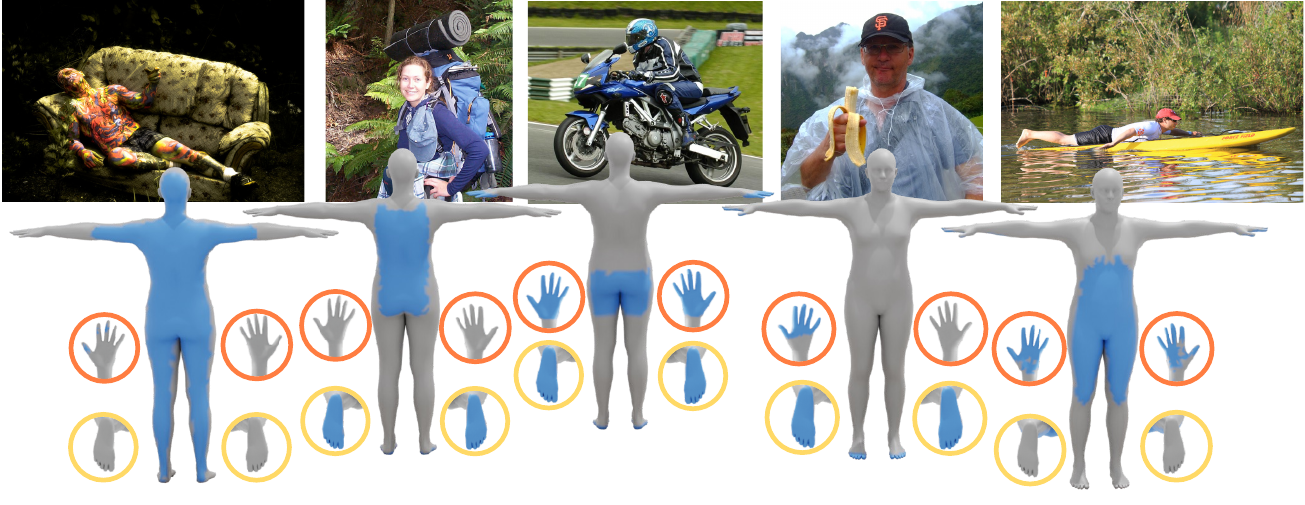}
    \vspace{-2.0 em}
    \caption{Given an RGB image, \nameMethod infers dense vertex-level \threeD contacts %
    on the full human body. 
    To this end, it 
    reasons about the contacting body parts, human-object proximity, and the surrounding scene context to infer \threeD contact for diverse human-object and human-scene interactions. 
    \textbf{\color{ContactBlue}{Blue areas}} show the inferred contact on the body, hands, and feet for each image.
    }
    \label{figure:teaser}
    \vspace{+1.5 em}
\end{center}%

}]

\begin{abstract}
\vspace{-1.0 em}
\renewcommand{\thefootnote}{\fnsymbol{footnote}}
\footnotetext[1]{equal technical contribution}
\footnotetext[2]{project lead}
Understanding how humans use physical contact to interact with the world is key to enabling human-centric artificial intelligence. 
While inferring \threeD contact is crucial for modeling realistic and physically-plausible human-object interactions, existing methods either focus on 2D, consider body joints rather than the surface, use coarse 3D body regions, or do not generalize to in-the-wild images.
In contrast, we focus on inferring dense, \threeD contact
between the full body surface and objects in arbitrary images.
To achieve this, we first collect \nameDataset, a new dataset containing dense vertex-level contact annotations paired with RGB images containing complex human-object and human-scene contact. 
Second, we train \nameMethod, a novel \threeD contact detector that uses both body-part-driven and scene-context-driven attention to estimate vertex-level contact on the \smpl body. 
\nameMethod builds on the insight that human observers recognize contact by reasoning about the contacting body parts, their proximity to scene objects, and the surrounding scene context. 
We perform extensive evaluations of our detector on \nameDataset as well as on the RICH and BEHAVE datasets. 
We significantly outperform existing SOTA methods across all benchmarks. 
We also show qualitatively that \nameMethod generalizes well to diverse and challenging real-world human interactions in natural images. 
The code, data, and models
\camready{are available at} \projectURL. 
\end{abstract}

\vspace{-1.0 em}

\section{Introduction}
\label{sec:intro}
Humans rely on contact to interact with the %
world. %
While we use our hands and feet to support grasping and locomotion, we also leverage our entire body surface in our daily interactions with the world; see \cref{figure:teaser}.
We sit on our buttocks and thighs, lie on our backs,
kneel on our knees, carry bags on our shoulders, 
and move heavy objects by holding them against our bodies.
Executing everyday tasks 
involves diverse full-body and object contact. 
Thus, modeling and inferring %
contact from images or videos is essential for %
applications such as human activity understanding, 
robotics, biomechanics, and augmented or virtual reality.

Inferring contact from images has recently received attention.
While some methods infer contact for hands \cite{narasimhaswamy2020detecting}, feet \cite{rempe2021humor}, self contact \cite{Mueller:CVPR:2021, Fieraru_2021_AAAI}, or person-person contact \cite{fieraru2020humanhuman}, 
others focus on human-scene or human-object contact for the full body \cite{HOT,RICH}.
HOT \cite{HOT} infers contact in \twoD by training on in-the-wild images with crowd-sourced \twoD contact areas, while BSTRO \cite{RICH} infers \threeD contact on a body mesh and is trained on images paired with \threeD body and scene meshes reconstructed with a multi-camera system.

\begin{figure}
\centerline{
\includegraphics[trim=000mm 001mm 000mm 002mm, clip=true, width=1.0\linewidth]{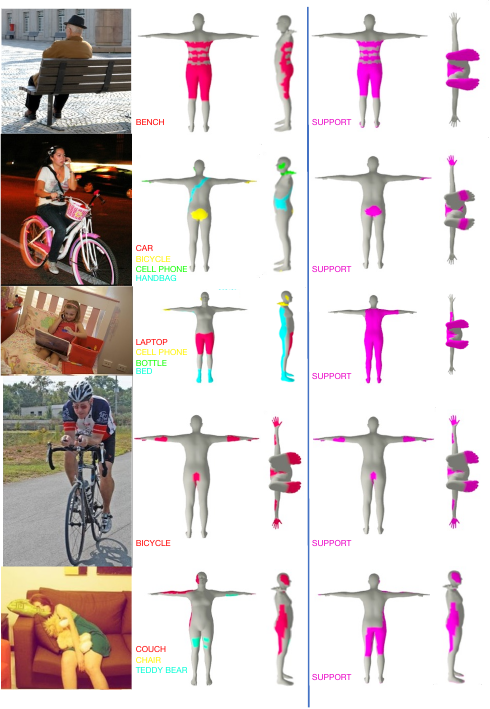}}
\vspace{-0.1in}
\caption{Sample contact annotations from the \nameDataset dataset. \textbf{Left to Right:} RGB image, two views showing human-supported contact (color-coded by object labels), and two views showing scene-supported contact.}
\label{figure:dataset_figure}
\vspace{-1.0 em}
\end{figure}

In contrast to prior work, we seek to represent detailed scene contacts across the full body and to infer these from in-the-wild images as illustrated in \cref{figure:teaser}.
To that end, we need both an appropriate training dataset and an inference method.
Note that manipulating objects is fundamentally \threeD. 
Thus, we must capture, model, and  
understand contact in \threeD. 
Also note that some contacts support the body, while others do not.
When sitting on a chair and drinking a cup of coffee, the body is supported by the buttocks on the chair and feet on the floor, while the coffee cup does not support the body.
The former is critical for physical reasoning about human pose and motion, while the latter is important to understand how we interact with objects. The \textit{type} of contact is therefore 
 important to represent. %
For a method to robustly estimate contact for arbitrary images we need a rich dataset that combines in-the-wild images with precise \threeD annotations; see \cref{figure:dataset_figure}.
This is a huge challenge.

To address this challenge, we present a novel method and a new dataset.
We first  %
collect 
a 
dataset with \threeD contact annotations for in-the-wild images using a novel interactive \threeD labelling tool (\cref{figure:dataset_figure}).
We then train 
a novel \threeD contact detector that takes a single image as input and produces dense contact labels on a \threeD body mesh (\cref{figure:teaser}).
Training on our new dataset means that the method generalizes well.

\zheading{Contact data}
To train a \threeD contact detector that is both 
accurate and robust, we need appropriate training data. 
However, existing datasets for \threeD contact~\cite{Hassan2019prox,bhatnagar2022behave,RICH} 
involve pre-scanning a \threeD scene and estimating \threeD human pose and shape (\HPS) of people  in the scene.
These approaches are limited in the complexity of the human-scene interactions, the size of the dataset, and very few methods capture human-object interactions paired with image data \cite{huang2022intercap,Brahmbhatt_2020_ECCV}.
An alternative is to use synthetic data \cite{shimada2022hulc}, but getting realistic synthetic data of complex human contacts is challenging, causing a domain gap between the dataset and real images.

In contrast, crowdsourced image annotations support many tasks in computer vision
such as 
image classification \cite{imageNet}, 
object detection~\cite{Lin2014MicrosoftCC, wu2019detectron2}, 
semantic segmentation~\cite{Lin2014MicrosoftCC,maskRCNN},  
\twoD human pose estimation~\cite{cao2019openpose, poseTrack}, and \threeD body shape estimation \cite{SHAPY,BodyTalk}. 
HOT \cite{HOT} takes this approach for human-object contact, but the labels are all in \twoD, while contact is fundamentally \threeD.
Consequently, we collect 
a large  dataset with dense \threeD contact annotations for in-the-wild images, called \nameDataset (\nameDatasetFull). 
We enable this with a new interactive software tool that lets people ``paint'' contact areas on a \threeD body mesh such that these reflect the observed contact in images. 
We use Amazon Mechanical Turk, %
train human annotators for our task, and collect a rich corpus of \threeD contact %
annotations for standard datasets of %
in-the-wild images of diverse human-object 
interactions, \ie, 
\vcoco~\cite{gupta2015vcoco} and 
\hake~\cite{li2020hake}; 
\cref{figure:dataset_figure} shows samples of our dataset. 
Note how contact and support regions are distinguished as are the semantic labels related to object contact.

\zheading{Contact detection}
As noted in the literature \cite{HOT,RICH}, contact areas are ipso facto occluded in images, thus, detecting contact requires reasoning about the involved body-parts and scene elements. 
To this end, BSTRO~\cite{RICH} uses a transformer \cite{lin2021end-to-end} with positional encoding based on body-vertex positions to implicitly learn the context around these, but has no explicit attention over body or scene parts. 
HOT~\cite{HOT,RICH}, on the other hand, focuses only on \twoD, 
pulls image features, and processes them with two branches in parallel, a contact branch and a body-part attention branch; the latter helps the contact features attend areas on and around body parts. 

We go beyond prior work to estimate detailed \threeD contact on the body.
Our method, \nameMethod (\nameMethodLong), introduces two technical novelties: 
(1) \nameMethod uses not only \emph{body-part-driven attention}, but also adds \emph{scene-context-driven attention}, as well as a \emph{cross-attention} module; %
this explicitly encourages contact features computed from the image to attend to meaningful areas both on (and near) body parts and scene elements. 
(2) 
\nameMethod uses a new \twoD Pixel Anchoring Loss (PAL) that relates the inferred \threeD contacts to the respective image pixels.
For this, 
we infer a \threeD body mesh with CLIFF \cite{li2022cliff} (SOTA for \HPS), detect which vertices of this are in contact with \nameMethod, project the \threeD contact vertices onto the image, and encourage them to lie in HOT's corresponding \twoD contact-area annotations. 
Note that this brings together both crowd-sourced \twoD and \threeD contact annotations. 

\zheading{Experiments}
We perform detailed quantitative experiments and find that \nameMethod outperforms BSTRO on the test sets of RICH and \nameDataset, when both are trained on the same data.
Ablation studies show that our two-branch architecture effectively combines body part and scene information.
We also provide ablation studies of the backbone and training data. \camready{We show that the inferred contact from \nameMethod significantly outperforms 
 methods that compute the geometric vertex distance between a reconstructed object and human mesh
\cite{zhang2020phosa, xie2022chore}.}
Finally, we use \nameMethod's estimated contact in the task of \threeD human pose and shape estimation and find that exploiting 
estimated contact 
improves accuracy.

\zheading{Contributions}
In summary, our contributions are
(1) 
We collect \nameDataset, a large-scale dataset with dense vertex-level \threeD contact annotations for %
in-the-wild images of %
human-object interactions.
(2) 
Using 
\nameDataset, 
we train \nameMethod, a novel regressor that cross-attends to both body parts and scene elements to predict \threeD contact on a body. 
\camready{\nameMethod outperforms existing contact detectors, and all its components contribute to performance.}
\camready{This shows that learning %
\threeD contact estimation from natural images is possible.} 
(3) 
We integrate \nameMethod's inferred \threeD contacts into a \threeD \HPS method and show that this boosts accuracy.
\camready{(4)} 
Our data, models, and code 
\camready{are available} at \projectURL.

\section{Related Work}
\label{sec:related_work}

\subsection{\twoD contact in images}

There exist multiple ways of representing human-object interactions (\hoi) and human-scene interactions (\hsi) in \twoD. Several \hoi methods~\cite{qi2018learning, wang2019deep, xu2019learning, kim2021hotr, zou2021_hoitrans} localize humans and objects as bounding boxes and assign a semantic label to indicate the \emph{interactions} between them. However, the interaction labels focus on action and do not support contact inference.
Chen~\etal~\cite{HOT} output image-aligned contact heatmaps and body-part labels directly from the RGB image by training a regressor on approximate \twoD polygon-level contact annotations. Some approaches learn part-specific contact regressors for hand~\cite{narasimhaswamy2020detecting, shan2020understanding} and foot~\cite{RempeContactDynamics2020} contact but only detect rough bounding boxes around contacting regions or joint-level labels. %
Such coarse image-based contact annotations are ambiguous and not sufficient for many downstream tasks.
We address these limitations by collecting a large-scale dataset of paired images and accurate vertex-level contact annotations directly on the \threeD \smpl mesh.    

Several methods estimate properties related to contact such as 
affordances~\cite{wang2017binge, roy2016multi, koppula2014physically}, contact forces~\cite{zhu2016inferring, yuan2021simpoe, Shimada2020PhysCapTOG} and pressure~\cite{scott2020image, grady2022pressurevision, funk2018learning}. However, collecting large datasets with ground-truth object affordances, forces, or pressure is challenging. 
Clever~\etal~\cite{Clever2020bodiesRest} use simulation and a virtual pressure mat to generate synthetic pressure data for lying poses.
Tripathi~\etal~\cite{tripathi2023ipman} exploit interpenetration of the body mesh with the ground plane as a heuristic for pressure. 
Recent work~\cite{yuan2021simpoe, Shimada2020PhysCapTOG, gartner2022diffphy} uses a physics simulator to infer contact forces. In contrast, we focus on annotating and estimating \threeD contact, which is universal in \hoi and is intuitively understood by annotators.

\subsection{Joint- \& patch-level \threeD contact}

\textbf{Joint-level contact.}
\threeD contact information is useful for \threeD human pose estimation~\cite{RempeContactDynamics2020, Shimada2020PhysCapTOG, xie2022chore}, \threeD hand pose estimation~\cite{Cao2021ReconstructingHI, Grady2021ContactOptOC, hasson2019obman}, \threeD body motion generation~\cite{taheri2021goal, rempe2021humor, Zhang_2021_LEMO, Zhang2020Generating3P, Zhang2020PLACEPL} and \threeD scene layout estimation~\cite{yi2022mime}. \threeD pose estimation approaches use joint-level contact to \emph{ground} the estimated \threeD human mesh~\cite{YamamotoCVPR2000, Fieraru_2021_NEURIPS, Hassan2019prox, zanfir2018monocular, zhang2020phosa} or encourage realistic foot-ground contact to avoid foot-skating artefacts~\cite{Ikemoto2006, rempe2021humor, shi2020motionet, Zhang_2021_LEMO, zou2020reducing}.   PhysCap~\cite{Shimada2020PhysCapTOG} and others~\cite{zanfir2018monocular, zou2020reducing, RempeContactDynamics2020, rempe2021humor} constrain the human pose by predicting skeleton joint-level foot-ground contact from video. Several approaches predict \threeD contact states of 2D foot joints detected from RGB images by manually annotating contact labels~\cite{zou2020reducing} or computing contact labels from \mocap datasets~\cite{RempeContactDynamics2020, Shimada2020PhysCapTOG}. Rempe~\etal~\cite{rempe2021humor} extend joint-level contact estimation to the toe, heel, knee and hands, but  use heuristics such as a zero-velocity constraint to estimate contact from AMASS~\cite{AMASS_2019}. Zhang~\etal~\cite{Zhang_2021_LEMO} estimate contact between foot-ground vertices using alignment of normals between foot and scene surface points. 
Such joint-level annotations cannot represent the richness of how human bodies contact the world.
In contrast \nameMethod captures dense vertex-level contact across the full body.

\textbf{Discrete patch-level contact.} Pre-defined contact regions or ``patches" on the \threeD body provide an intermediate representation for modeling surface-level contact. M\"uller~\etal~\cite{Mueller:CVPR:2021} and Fieraru~\etal~\cite{Fieraru_2021_AAAI} crowdsource patch-level self-contact annotations between discrete body-parts patches on the same individual. Fieraru~\etal~\cite{fieraru2020humanhuman} also collect  patch-level contact between two interacting people.
While richer than joint-level contact,  \emph{patches} 
do not model fine-grained contact.
In contrast, the \nameDataset dataset and \nameMethod model contact on the vertex level, significantly increasing the contact resolution.   

\subsection{Dense vertex-level contact }
Dense ground-truth contact can be computed if one has accurate \threeD bodies in \threeD scenes.
For instance, PROX~\cite{Hassan2019prox}, InterCap~\cite{huang2022intercap}, and BEHAVE~\cite{bhatnagar2022behave} use RGB-D cameras to capture humans interacting with objects and scenes whereas HPS~\cite{guzov2021hps} uses a head-mounted camera and IMU data to localize a person in a pre-scanned \threeD scene. RICH uses a laser scanner to capture high-quality \threeD scenes and the bodies are reconstructed using multi-view cameras. GRAB~\cite{GRAB:2020} captures hand-object interactions using marker-based \mocap but lacks images paired with the ground-truth scene. 
Such datasets require a constrained capture setup and are difficult to scale. 
An alternative uses synthetic \threeD data.
HULC~\cite{shimada2022hulc} generates contact by fitting SMPL  to 3D joint trajectories in the GTA-IM~\cite{cao2020gta-im} dataset. 
The contacts, however, lack detail and the domain gap between the video game and the real world limits generalization to natural images.

Several methods infer \threeD bodies using dense \threeD contact.
PHOSA~\cite{zhang2020phosa} jointly estimates \threeD humans, objects and contacts for  a limited set of objects for which there are predetermined, hand-crafted, contact pairs on the human and object. 
Other methods optimize the body and scene together using information about body-scene contact \cite{weng2020holistic,yi2022mime,xie2022chore,xie2023vistracker,bite2023rueegg}.

Some methods predict dense contact on the body mesh.
POSA~\cite{Hassan2021posa} learns a body-centric prior over  contact.  Given a posed \threeD body, POSA predicts which vertices are likely to contact the world and what they are likely to contact.
It assumes the pose is given.
Closest to our work are BSTRO~\cite{RICH} and HULC~\cite{shimada2022hulc}, which infer dense contact on the body from an image.
We go beyond these methods by providing a rich dataset of images in the wild with dense contact labels.
Moreover we exploit contextual cues from body parts as well as the scene and objects using a novel attentional architecture.

\section{\nameDataset Dataset}
\label{sec:dataset}
\nameDataset  is a collection of \emph{vertex-level} \threeD contact labels on \smpl paired with color images of people in unconstrained environments with a wide diversity of human-scene and human-object interactions. 
We source our images from the HOT dataset~\cite{HOT} for the following reasons: 
(1) 
HOT curates valid human contact images from existing \hoi datasets like V-COCO~\cite{gupta2015vcoco} and HAKE~\cite{li2020hake} by removing indirect human-object interactions, heavily cropped humans, motion blur, distortion or extreme lighting conditions; 
(2) 
HOT contains 15082 images containing \twoD \emph{image-level} contact annotations, which are complementary to the dense \threeD contact annotations in our dataset. 
Example images and contact annotations from the \nameDataset dataset are shown in \cref{figure:dataset_figure}.

\subsection{Types of contact }
While existing \hoi methods and datasets typically treat all contacts the same way,
human contact is more nuanced. 
Physical contact can be classified into 3 categories: 
(1) \textit{scene-supported contact}, \ie, humans supported by scene objects; 
(2) \textit{human-supported contact}, \ie, objects supported by a human; and 
(3) \textit{unsupported} contact, \eg,  
self-contact~\cite{Mueller:CVPR:2021, Fieraru_2021_AAAI} and human-human contact~\cite{Fieraru_2021_NEURIPS, fieraru2020humanhuman}. 
Since datasets for the latter already exist, %
we focus on the first two categories, \ie, contact that involves support. 
\camready{Note that labeling contact in images is challenging.  Focusing on support helps reduce ambiguous cases where humans are close to scene objects but not actually in contact. }
We use Amazon Mechanical Turk (AMT) to crowd-source annotations for \nameDataset; 
we ask 
people to annotate both \emph{human-supported contact} for each individual object and \emph{scene-supported contact}. 

\subsection{Annotation procedure}
\label{subsec:DAMON:annotation_procedure}

We create a novel user-friendly interface and tool that enables 
annotators 
to \emph{``paint''} \threeD vertex-level contact areas directly on the human mesh; see 
the interface %
in \supmat 
We show the original image with the type of contact to be annotated on the left and the human mesh to the right. 
We then ask 
annotators 
to ``paint'' contact labels on the $N_V = 6890$ vertices of the SMPL~\cite{SMPL:2015} template mesh, $\smplmesht \in \mathbb{R}^{6890\times3}$.

The tool has features such as mesh rotation, zoom in/out, paint-brush size selection, 
an eraser, and a reset button.
Depending on the selected brush size, the tool ``paints'' contact annotations by selecting a \emph{geodesic} neighborhood of vertices around the vertex currently under the mouse pointer. 
For a detailed description of the tool, see \video in \supmat

The tool lets annotators label contact with multiple objects in addition to the scene-supported contact. 
For example annotations, see \cref{figure:dataset_figure}. 
For every image, to label human-supported contact, we cycle through object labels provided in the 
\vcoco 
and 
\hake 
datasets. %
For scene-supported contact, we ask 
annotators 
to label contact with all supporting scene objects, including the ground. %
We automatically get body-part labels for contact vertices using \smpl's part segmentation. 
To support amodal contact estimation, we ask 
annotators 
to also label %
contact regions that may not be visible in the image but can be guessed confidently. 
We filter out ambiguous contact in images such as human-human contact, human-animal contact, and indirect human-object interactions, such as pointing; for details \camready{about data collection and how we limit ambiguity in the task}, see \supmat

\begin{figure}[t]
\centerline{\includegraphics[trim=70 150 200 200,clip,width=\linewidth]{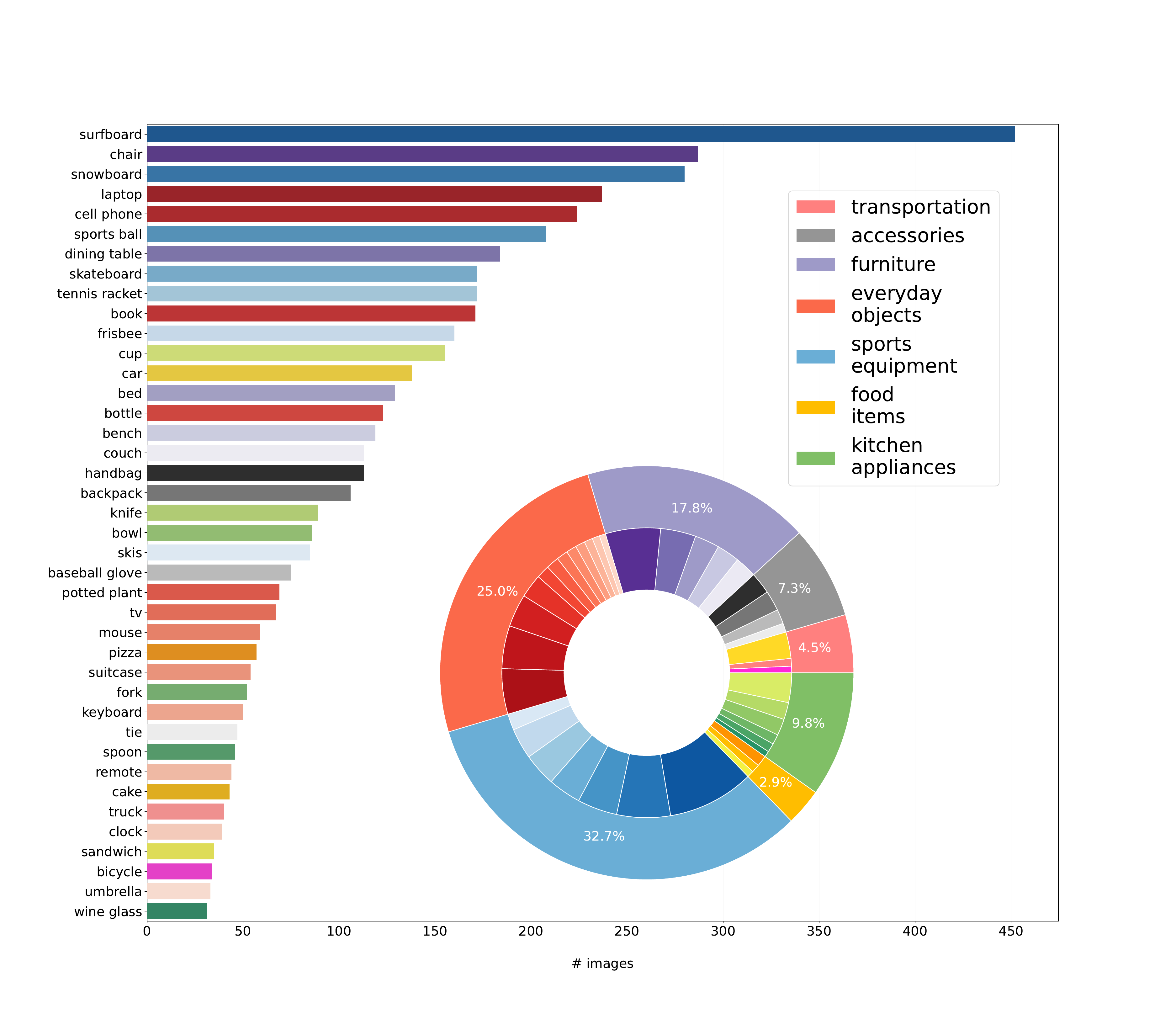}}
\caption{
            \nameDataset dataset statistics. %
            \textbf{Histogram}: 
            contact object labels ($y$-axis) and the number of images in which they are present ($x$-axis). 
            We crop the plot in the interest of space; for the full long-tailed plot see \supmat 
            \textbf{Pie chart}: 
            object labels are grouped into 7 main categories; 
            inner colors correspond to the colors in the histogram.
            \faSearch~\textbf{Zoom in}. %
}
\label{figure:damon:stats1}
\end{figure}

We ensure a high annotation quality with two quality checks:
(1) 
We detect and filter out the inconsistent annotators; 
out of 100 annotators we keep only 14 good ones. 
(2) 
We have meta-annotators curate the collected annotations; 
images with noisy annotations are then pushed for a re-annotation.
For details \camready{about quality control}, see \supmat

We access \nameDataset's quality by computing two metrics:
\mbox{(1)
\emph{Label accuracy:}} 
We manually curate from RICH~\cite{RICH} and PROX~\cite{Hassan2019prox} 100 images that have highly-accurate \threeD poses and contact labels. 
We treat these as ground-truth contact, and compute the IoU of our collected annotations. 
\mbox{(2)
\emph{Level of annotators' agreement:}}
We ask annotators to label the same set of 100 images, and compute \emph{Fleiss' Kappa ($\kappa$)} metric.
For a detailed analysis of results, see \supmat

\subsection{Dataset statistics}

Out of HOT's 15082 images we annotate 5522 images via our annotation tool (\cref{subsec:DAMON:annotation_procedure}); 
we ``paint'' contact vertices, and assign to each vertex an appropriate label out of 84 object (\cref{figure:damon:stats1}) and 24 body-part labels.  
An image has on average \threeD contacts for 1.5 object labels.
We use HOT's train/test/val data splits. 

\begin{figure}[ht]
\centering
\includegraphics[width=\linewidth]{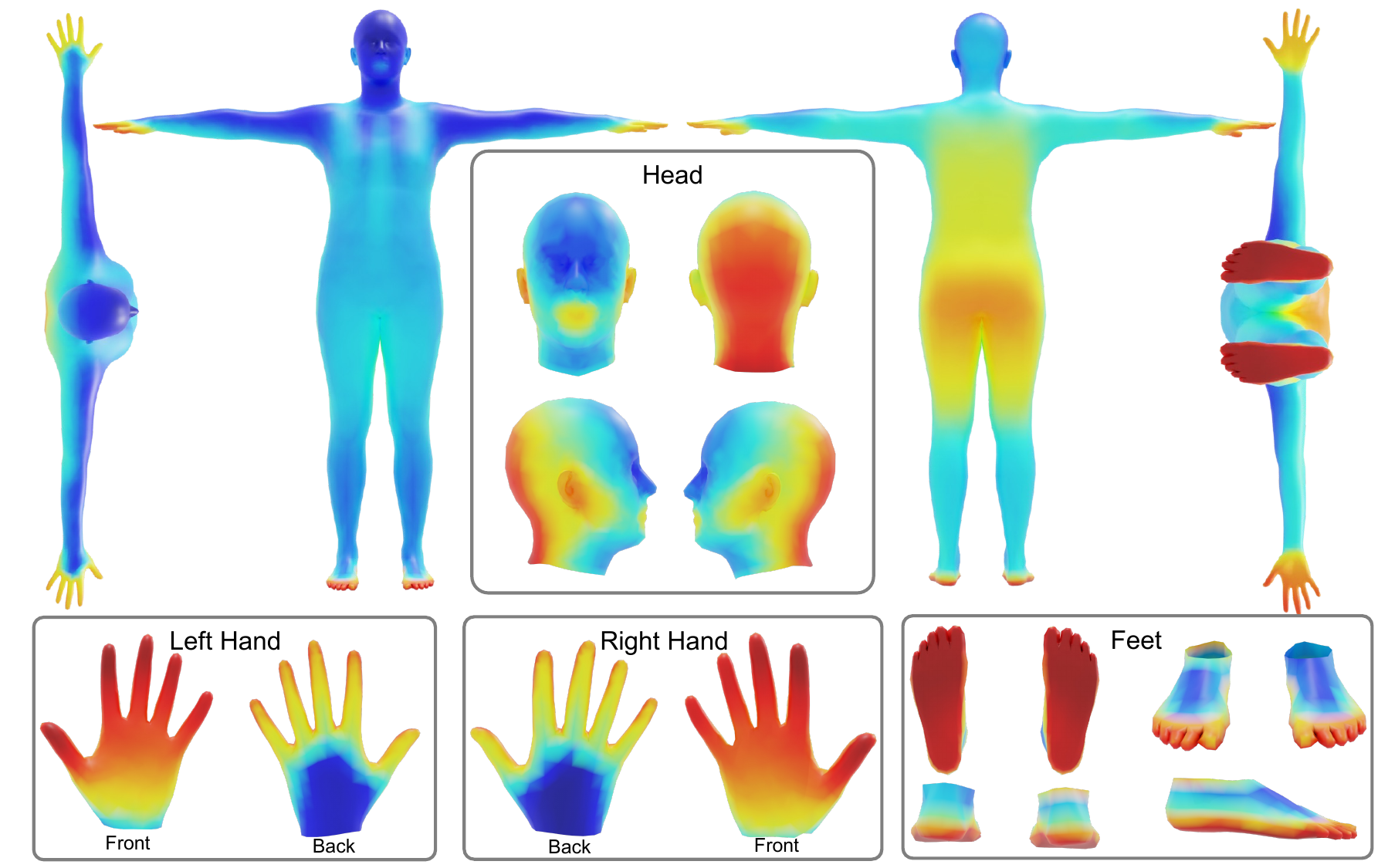}
\caption{Aggregate statistics showing contact probabilities across the body vertices in the \nameDataset dataset. The body part closeups show the contact probabilities normalized for that body part. \textcolor{red}{Red} implies higher probability of contact while \textcolor{blue}{blue} implies lower probability. \faSearch~\textbf{Zoom in}.}
\label{figure:supmat:contact_probability_mesh}
\end{figure}

\camready{We also show aggregate vertex-level contact probabilities on the \smpl mesh across the whole \nameDataset dataset in \cref{figure:supmat:contact_probability_mesh}. The individual body-part close-ups in \cref{figure:supmat:contact_probability_mesh} show normalized contact probabilities for that body part. It is evident that, while we typically use our hands and feet for contact, we also frequently use the rest of our body, especially the buttocks, back of the head, chest, lips, and ears to interact with everyday objects. 
To our knowledge, no such analysis of full-body contact for in-the-wild images has previously been reported.
This motivates the need for modeling dense full-body contact.}

\section{Method: \nameMethod}
\label{sec:method}

Contact regions in images are ipso facto occluded. 
This makes human-object contact estimation from in-the-wild images a challenging and ill-posed problem. 
We tackle this with a new \textbf{DE}nse \textbf{CO}ntact estimator, \nameMethod, which uses scene and part context. 

Our contributions are two fold: 
(1) 
To reason about the contacting body parts, human-object proximity, and the surrounding scene context, we use a novel architecture with three branches, \ie, 
a scene-context, 
a part-context, and 
a per-vertex contact-classification branch. 
(2) 
We use a novel \twoD pixel-anchoring loss that constrains the solution space by grounding the inferred \threeD contact to the \twoD image space.       

\subsection{Model architecture} \label{sec:architecture}

\begin{figure*}[ht]
\centering
\includegraphics[width=\linewidth]{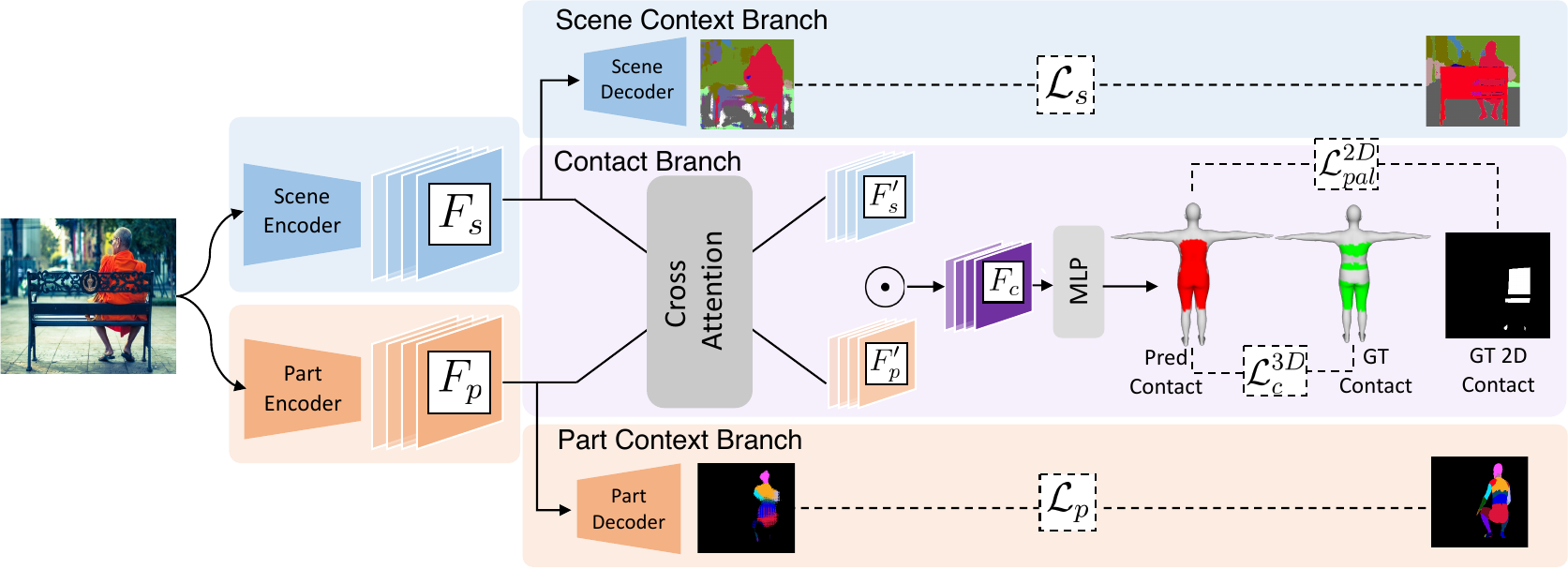}
\vspace{-1.5 em}
\caption{
            \nameMethod architecture (\cref{sec:architecture}). 
            \nameMethod reasons about body parts, human-object proximity, and the surrounding scene context. 
            To this end, it uses three branches, \ie, a scene-context, a part-context, and a per-vertex contact-classification branch.
            Cross attention guides the features to focus attention %
            on (and around) body parts and scene elements that are relevant for contact. 
}
\label{figure:deco_architecture}
\end{figure*}

Given an image $\mathbf{I} \in \mathbb{R}^{H \times W \times 3}$, \nameMethod predicts contact probabilities on the \smpl~\cite{SMPL:2015} mesh. 
We use \smpl as it is widely used for \HPS estimation 
\cite{kolotouros2019spin, Kocabas2021pare, Khirodkar2022OccludedHM, li2022cliff, Kanazawa2018_hmr, Zhang2021PyMAF3H}. \smpl parameterizes the human body with pose and shape parameters, $\boldsymbol{\Theta} = [\boldsymbol{\theta} \in \mathbb{R}^{72}, 
\boldsymbol{\beta} \in \mathbb{R}^{10}]$ and outputs a \threeD mesh $\smplmesh(\boldsymbol{\theta}, \boldsymbol{\beta})\in\mathbb{R}^{6890\times3}$. 
\smpl's template mesh $\smplmesht$ is segmented 
into $J = 24$ parts, $\meshPartk \in \meshPartSET$, which allows part-labeling of contact vertices. 
Moreover, \smpl's mesh topology is consistent with the \smplh~\cite{romero2017smplh} model and has the same vertices below the neck as the \smplx model~\cite{RICH}, making our contact representation widely applicable. 

\Cref{figure:deco_architecture} shows \nameMethod's architecture. 
Intuitively, contact estimation relies on both part and scene features as they are complementary. We use two separate encoders $\mathcal{E}_s$ and $\mathcal{E}_p$ to extract scene features $\boldsymbol{F_s}$ and body-part features $\fpart$. For the encoder backbone, we use both the transformer-based SWIN~\cite{liu2021swin} and the CNN-based HRNET~\cite{wang2021hrnet}.
We integrate scene features $\fscene$ and body-part features $\boldsymbol{F_p}$ via a cross-attention module inspired by \cite{vaswani2017transformer, lu2019vilbert}. 
Previous methods either concatenate multi-modal features~\cite{mehta2017vnect}, 
use channel-wise multiplication~\cite{Kocabas2021pare}, adopt trainable fusion~\cite{tekin2017fusion} or use bilinear interpolation between multi-modal features~\cite{sun2019bilinear}. 
However, such methods simply combine the multi-modal features without explicitly exploiting their interactions. 
In contrast, \nameMethod's cross-attention guides the network to ``attend'' to relevant regions in $\boldsymbol{F_s}$ and $\boldsymbol{F_p}$ to reason about contact.  

To implement cross-attention, we exchange the key-value pairs in the multi-head attention block between the two branches. Specifically, we 
\camready{initialize}
the query, key, and value matrices for each branch \ie $\{\mathcal{Q}_s, \mathcal{K}_s, \mathcal{V}_s\}=\{\fscene, \fscene, \fscene\}$ for the scene branch and $\{\mathcal{Q}_p, \mathcal{K}_p, \mathcal{V}_p\}=\{\fpart, \fpart, \fpart\}$ for the part branch. Then we obtain the contact features $\fcontact$ after multi-head attention as 
\begin{align}
    \fscene' &= \text{softmax}(\mathcal{Q}_p\mathcal{K}_s^T/\sqrt{C_t})\mathcal{V}_s    \text{,}    \\
    \fpart' &= \text{softmax}(\mathcal{Q}_s\mathcal{K}_p^T/\sqrt{C_t})\mathcal{V}_p     \text{,}    \\
    \fcontact &= \mathit{LN}(\fscene' \odot \fpart')                                    \text{,}
\end{align}
where $C_t$ is a scaling factor~\cite{vaswani2017transformer}, $\odot$ is the Hadamard operator and $\mathit{LN}$ represents layer-normalization~\cite{ba2016layernorm}. We obtain final contact predictions $\bar{y}_c \in \mathbb{R}^{6890\times1}$ after filtering $\fcontact$ via a shallow MLP followed by sigmoid activation.  

The \nameMethod architecture encourages the scene and part encoders, $\mathcal{E}_s$ and $\mathcal{E}_p$, to focus on relevant features by upsampling $\fscene$ and $\fpart$ using scene decoder $\mathcal{D}_s$ and part decoder $\mathcal{D}_p$ respectively. The output of $\mathcal{D}_s$ is a predicted scene segmentation map, $\boldsymbol{\bar{X}}_s \in \mathbb{R}^{H\times W \times N_o}$, where $N_o$ are the number of objects in MS COCO~\cite{lin2014coco}. Similarly, we obtain the part features $\boldsymbol{\bar{X}}_p \in \mathbb{R}^{H\times W \times (J+1)}$ from $\mathcal{D}_p$, where $J$ are the number of body parts and the extra channel is for the background class. 

We train \nameMethod end-to-end (\cref{figure:deco_architecture}) with the loss: %
\begin{align}
    \mathcal{L} = w_c\mathcal{L}_c^{3D} + w_{pal}\mathcal{L}_{pal}^{2D} + w_s\mathcal{L}_s^{2D} + w_p\mathcal{L}_p^{2D}
    \text{,}
\end{align}
where $\mathcal{L}_c^{3D}$ is the binary-cross entropy loss between per-vertex predicted contact $\bar{y}_c$ and ground-truth contact labels $y^{gt}_c$. $\mathcal{L}_s^{2D}$ and $\mathcal{L}_p^{2D}$ are segmentation losses between the predicted and the ground-truth masks. We describe $\mathcal{L}_{pal}^{2D}$ in the following section. 
Steering weights $w$ are set empirically.

\subsection{\twoD Pixel Anchoring Loss (PAL)}

\begin{figure}[t]
\centering
\includegraphics[width=\linewidth]{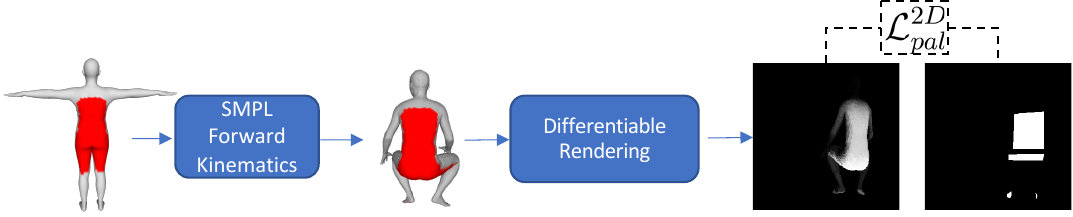}
\caption{   
            The Pixel Anchoring Loss (PAL) grounds \threeD contact predictions to image pixels by rendering the contact-colored posed mesh on the image plane. 
            The rendered contact mask is compared with \twoD contact ground truth contact from HOT~\cite{HOT}}.
\label{figure:pal_loss}
\end{figure}

To relate contact on the \threeD mesh with image pixels, we propose a novel pixel anchoring loss (PAL); see \cref{figure:pal_loss}.
We run the SOTA \hps network CLIFF~\cite{li2022cliff} on input image $I$ to infer the camera scale $s$, camera translation, $\mathbf{t}^c$, and \smpl parameters, $\pose$ and $\shape$, in the camera coordinates assuming camera rotation,
$\camrot=\bs{I}_3$ and body translation, 
$\bodytransl=\mathbf{0}$. Using the estimated \smpl parameters, we obtain the \emph{posed} mesh $\mathcal{M}(\pose, \shape, \bodytransl)$, which is colored using \nameMethod-predicted per-vertex contact probability, $\bar{y}_c$, in a continuous and differentiable manner. We denote the posed mesh colored with contact probability by $\mathcal{M}_c$.
We use the PyTorch3D~\cite{ravi2020pytorch3d} differentiable renderer 
to render $\mathcal{M}_c$ on the image under weak perspective, resulting in the \twoD contact probability map, $\boldsymbol{\bar{X}}_c^{2D}$. $\mathcal{L}^{2D}_\text{pal}$ is computed as the binary-cross entropy loss between $\boldsymbol{\bar{X}}_c^{2D}$ and the ground-truth \twoD contact mask from HOT~\cite{HOT}, $\boldsymbol{X}_c^{2D}$.

\section{Experiments}
\label{sec:experiments}
\textbf{Implementation Details.}
We experiment with both Swin Transformer~\cite{liu2021swin} and HRNET~\cite{wang2021hrnet} as backbone architectures for $\mathcal{E}_s$ and $\mathcal{E}_p$. 
We initialize the two encoder configurations with ImageNet and HRNET pretrained weights respectively. 
We obtain pseudo ground-truth scene segmentation masks, $\boldsymbol{X}_s \in \mathbb{R}^{H\times W \times N_o}$, containing semantic labels for $N_o=133$ categories, by running inference using the SOTA image segmentation network, Mask2Former~\cite{mask2former}. 
To get ground-truth part segmentations, $\boldsymbol{\bar{X}}_p \in \mathbb{R}^{H\times W \times (J+1)}$, we follow \cite{Kocabas2021pare} to use the \smpl part segmentation and segment the posed ground-truth mesh when available (\eg in RICH and PROX) into $J=24$ parts, rendering each part mask as a separate channel. 
Since there are no ground-truth \threeD meshes in \nameDataset, we obtain pseudo ground-truth meshes by running the SOTA human pose and shape network, CLIFF~\cite{li2022cliff}. 
\camready{This strategy works better in practice than using a human-parsing network (\eg Graphonomy~\cite{gong2019graphonomy}). It has the advantage of \emph{left-right sided} part labels, which helps in circumventing left-right ambiguity. It also retains full-visibility under occlusion, which allows reasoning about parts not visible in the original image.}

\textbf{Training and Evaluation.}
To train \nameMethod, we use the \nameDataset dataset along with existing datasets with \threeD contact labels: RICH~\cite{RICH} and PROX~\cite{Hassan2019prox}. 
We evaluate our method on the test splits of \nameDataset and RICH. 
To evaluate out-of-domain generalization performance, we also show evaluation on the test split of  BEHAVE~\cite{bhatnagar2022behave}, which is not used in training. 
We follow \cite{RICH} and report both count-based evaluation metrics: precision, recall and F1 score and geodesic error (in cm, see \cite{RICH} for details). 
For additional implementation and training details, please refer to \supmat

\subsection{\threeD Contact Estimation} 
\label{sec:evaluationContactEst}

We compare \nameMethod with BSTRO~\cite{RICH} and POSA~\cite{Hassan2021posa}, both of which give  dense vertex-level contact on the body mesh.
Since POSA needs a posed body mesh as input, we show POSA results when given  ground-truth meshes, called POSA$^\text{GT}$ and meshes reconstructed by PIXIE~\cite{feng2021pixie}, called POSA$^\text{PIXIE}$. 
For a fair comparison, we make sure to use the same training data splits in all our evaluations. 

We report results on RICH-test, BEHAVE-test, and \nameDataset-test in \cref{table:compare_sota}. 
For evaluation on RICH-test, we train both BSTRO and \nameMethod on the RICH training split only. 
This ablates the effect of the \nameDataset dataset, allowing us to isolate the contribution of the \nameMethod architecture. 
As shown in \cref{table:compare_sota}, we outperform all baselines across all metrics. 
Specifically, we report a significant $\sim$11\% improvement in F1 score and 7.93 cm improvement in the geodesic error over the closest baseline, BSTRO. 
Further, we observe that adding $\mathcal{L}_{pal}^{2D}$ improves the geodesic error considerably with only a slight trade-off in F1 score. 
Here, we reiterate the observation in \cite{RICH} that, while POSA matches \nameMethod in recall, it comes at the cost of precision, resulting in worse F1 scores. 
Since POSA does not rely on image evidence and only takes the body pose as input, it tends to predict false positives.
For qualitative results, see \cref{figure:deco_qual_comparison} and \supmat

\begin{table*}[!htbp]
\centering
\resizebox{\textwidth}{!}{
\begin{tabular}{l|cccc|cccc|cccc}
\Xhline{3\arrayrulewidth}
\multirow{2}{4em}{\bf Methods} & \multicolumn{4}{c|}{\bf RICH~\cite{RICH}} & \multicolumn{4}{c|}{\bf \nameDataset} & \multicolumn{4}{c}{\bf BEHAVE~\cite{bhatnagar2022behave}} \\ 
\cline{2-13}
 & \bf Precision $\uparrow$ & \bf Recall $\uparrow$ & \bf F1 $\uparrow$ & \bf geo.  (cm)$\downarrow$ & \bf Precision $\uparrow$ & \bf Recall $\uparrow$ & \bf F1 $\uparrow$ & \bf geo.  (cm) $\downarrow$ & \bf Precision $\uparrow$ & \bf Recall $\uparrow$ & \bf F1 $\uparrow$ & \bf geo. (cm) $\downarrow$ \\ \hline
BSTRO~\cite{RICH} & 0.65 & 0.66 & 0.63 & 18.39 & 0.51 & 0.53 & 0.46 & 38.06 & 0.13 & 0.03 & 0.04 & 50.45 \\ \hline
POSA$^{\text{PIXIE}}$~\cite{feng2021pixie, Hassan2021posa} & 0.31 & 0.69 & 0.39 & 21.16 & 0.42 & 0.34 & 0.31 & 33.00 & 0.11 & 0.07 & 0.06 & 54.29 \\ \hline
POSA$^{\text{GT}}$~\cite{feng2021pixie, Hassan2021posa} & 0.37 & \bf 0.76 & 0.46 & 19.96 & - & - & - & - & 0.10 & 0.09 & 0.06 & 55.43 \\ \hline
\hline
\nameMethod & \bf 0.71 & \bf 0.76 & \bf 0.70 & 17.92 & 0.64 & \bf 0.57 & \bf 0.55 & \bf 21.32 & 0.25 & \bf 0.21 & \bf 0.18 & 46.33 \\ \hline
\nameMethod$+\mathcal{L}_{pal}^{2D}$ & \bf 0.71 & 0.74 & 0.69 & \bf 10.46 & \bf 0.65 & \bf 0.57 & \bf 0.55 & 21.88 & \bf 0.27 & 0.18 & \bf 0.18 & \bf 44.51 \\ 
\Xhline{3\arrayrulewidth}
\end{tabular}}
\vspace{-0.5 em}
\caption{
            Comparison of \nameMethod with SOTA models on RICH~\cite{RICH}, \nameDataset, and BEHAVE~\cite{bhatnagar2022behave}. 
            See discussion in \cref{sec:evaluationContactEst}. 
}
\label{table:compare_sota}
\end{table*}

Next, we retrain both BSTRO and \nameMethod on all available training datasets, RICH, PROX and \nameDataset, and evaluate on the \nameDataset test split. 
POSA training needs a GT body which is not available in \nameDataset. 
This evaluation tests generalization to unconstrained Internet images. 
Note that to train with $\mathcal{L}_{pal}^{2D}$, we include HOT images with \twoD contact annotations even if they do not have \threeD contact labels from \nameDataset. 
For these images, we simply turn off $\mathcal{L}_c^{3D}$. 
This is because \nameMethod, unlike BSTRO, is compatible with both \threeD and \twoD contact labels. 
\nameMethod significantly outperforms all baselines and results in an F1 score of 0.55 vs 0.46 for BSTRO with a 16.18 cm improvement in geodesic error. 
Notably, the improvement over baselines when including PROX and \nameDataset in training is higher compared with training only on RICH, which indicates that  \nameMethod scales better with more training images compared to BSTRO.  

Finally, we evaluate out-of-domain generalization on the unseen BEHAVE~\cite{bhatnagar2022behave} dataset. 
BEHAVE focuses on a single human-object contact per image, even if multiple contacting objects may be present. %
The focus on single object-contact in the GT contact annotations partly explains why most methods struggle with this dataset. 
Further, since BEHAVE does not label contact with the ground, for the purpose of evaluation, we mask out contact predictions on the feet. 
As reported in \cref{table:compare_sota}, we outperform all baselines on both F1 and geodesic error, which indicates that \nameMethod has a better generalization ability. %

\definecolor{ContactGreen}{rgb}{0.1, 0.8, 0.1}
\begin{figure*}%
\centering %
\includegraphics[trim=000 000 000 000,clip,width=\linewidth]{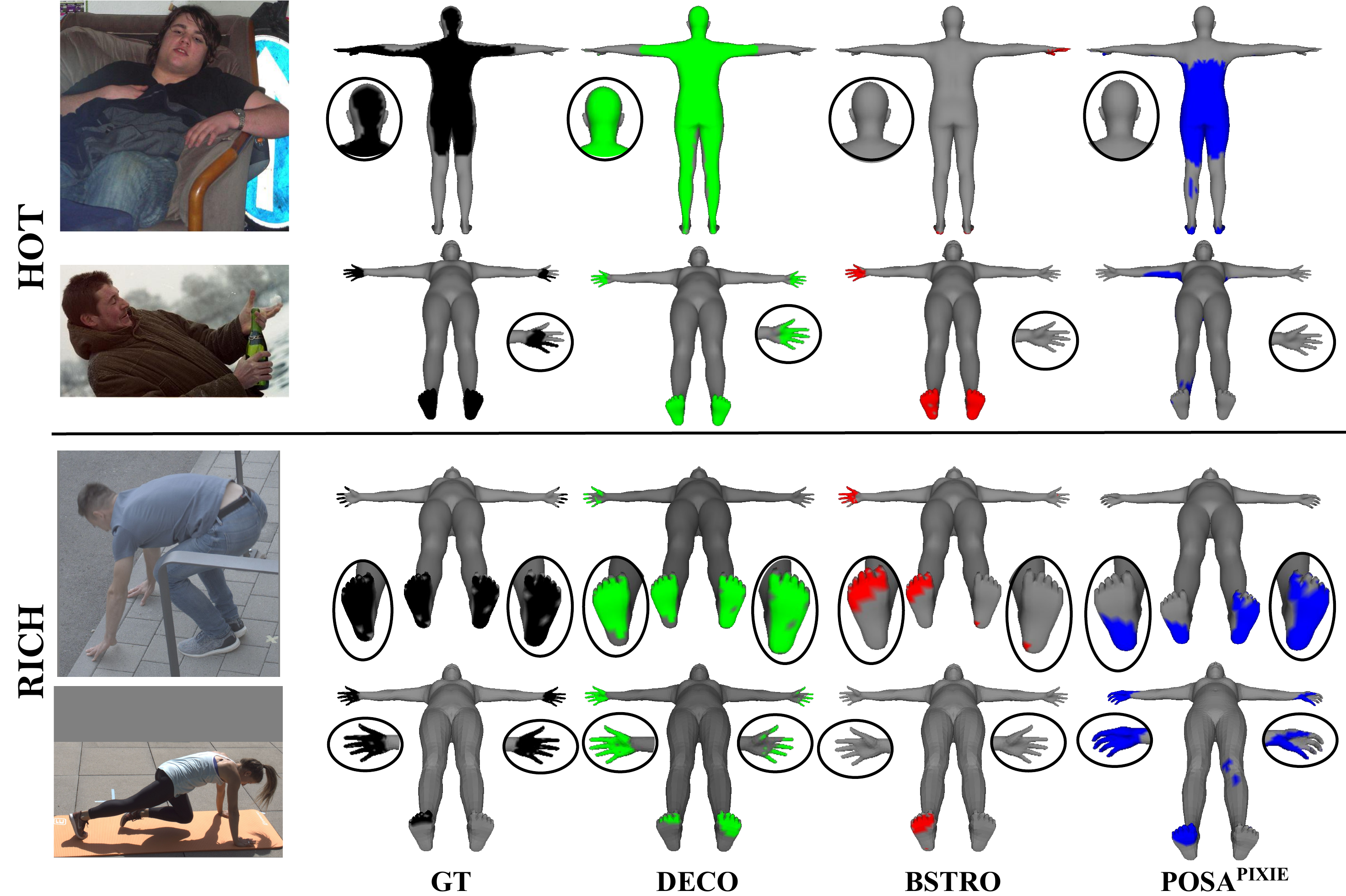}
\vspace{-1.5 em}
\caption{
            Qualitative evaluation of 
            \nameMethod             (\textcolor{ContactGreen}{green}), 
            BSTRO                   (\textcolor{red}{red}) and 
            POSA$^{\text{PIXIE}}$   (\textcolor{blue}{blue}), 
            alongside Ground Truth  (\textbf{black}). %
}
\label{figure:deco_qual_comparison}
\end{figure*}

\begin{table}[t] %
\centering
\resizebox{\columnwidth}{!}{
\begin{tabular}{cc|cc|c|cccc}
\Xhline{3\arrayrulewidth}
$\boldsymbol{\mathcal{E}_s}$ & $\boldsymbol{\mathcal{E}_p}$ & $\boldsymbol{\mathcal{L}_s^{2D}}$ & $\boldsymbol{\mathcal{L}_p^{2D}}$ & \bf Back. & \bf Pre. $\uparrow$ & \bf Rec. $\uparrow$ & \bf F1 $\uparrow$ & \bf geo. (cm) $\downarrow$ \\ 
\cline{1-9}
\multicolumn{2}{c|}{shared} & \xmark & \xmark  & HR & 0.68 & \bf 0.76 & 0.68 & 20.85 \\ \hline
\cmark & \cmark & \xmark & \xmark  & HR & 0.67 & 0.76 & 0.67 & 23.54 \\ \hline
\cmark & \cmark & \cmark & \xmark  & HR & 0.68 & 0.75 & 0.68 & 18.44 \\ \hline
\cmark & \cmark & \xmark & \cmark  & HR & 0.70 & 0.74 & 0.68 & 18.37 \\ \hline
\cmark & \cmark & \cmark  & \cmark & SW & 0.68 & 0.71 & 0.66 & 18.54 \\ \hline
\cmark & \cmark & \cmark & \cmark  & HR & \bf 0.71 & \bf 0.76 & \bf 0.70 & \bf 17.92 \\ 
\Xhline{3\arrayrulewidth}
\end{tabular}
}
\vspace{-0.5 em}
\caption{
            Ablation study for \nameMethod design choices (\cref{sec:ablation}).
            We ablate: 
            (1) using separate or joint encoders for the scene and body parts, 
            (2) using branch-specific losses, 
            (3) using an HRNET (HR) or Swin (SW) backbone. 
            Bold denotes best performance. 
}
\label{table:deco:model_ablation}
\end{table}

\subsection{Ablation Study} \label{sec:ablation}

In \cref{table:deco:model_ablation} we evaluate the impact of our design choices. 
First, we analyze the effect of using a shared encoder for the scene and the part branch vs separate encoders for both. 
Compared to having separate encoders without branch-specific losses, a single encoder performs better, which can be attributed to having fewer training parameters. 
However, any configuration using  ${\mathcal{L}_s^{2D}}$ or ${\mathcal{L}_p^{2D}}$ outperforms the shared encoder. 
While ${\mathcal{L}_p^{2D}}$ contributes improvements to precision, ${\mathcal{L}_s^{2D}}$ contributes to better recall. 
This is expected since, intuitively, attending to body parts helps with inferring fine-grained contact, whereas scene context helps to reason about the existence of contact regions. 
Each one separately helps with geodesic error, but the best performance comes when used together, in terms of both F1 score and geodesic error. 
Finally, we see that the HRNET backbone outperforms the Swin  backbone. 
This is likely because HRNET is pretrained on human-centric tasks (like our task), whereas Swin in pretrained on ImageNet image classification. 

\subsection{Inferred versus geometric contact}

\camready{
An alternative to directly inferring contact, as \nameMethod does, is to first recover the \threeD body and scene and then compute contact geometrically using the distance between the body and scene
\cite{zhang2020phosa, xie2022chore}. If \threeD human and scene recovery were accurate, this could be a viable alternative to \nameMethod's inferred contact. 
To test this %
hypothesis %
we perform an experiment using the 
two SOTA techniques for 3D human and object estimation, PHOSA~\cite{zhang2020phosa} and CHORE~\cite{xie2022chore}.
PHOSA works only on 8 objects, and CHORE works on 13.
In contrast, \nameMethod supports all 80 object classes in MS-COCO. 
Because they are optimization based, PHOSA and CHORE are slow, taking 4 mins and 66 secs per image respectively. \nameMethod is real-time and takes 0.012 secs for inference. 
For fair comparison, we split the \nameDataset dataset and evaluate using test sets that include only objects supported by either PHOSA or CHORE. 
We reconstruct the human and object and then recover contact using thresholded distance.
CHORE achieves an F1 score of 0.08 as opposed to \nameMethod's score of 0.48. 
Similarly, PHOSA achieves an F1 score of 0.18 as opposed to \nameMethod's score of 0.60.
Given the current state of 3D human pose and scene estimation, \nameMethod significantly outperforms geometry-based contact estimation. 
}

\section{\hps using DECO contacts}

\camready{
Next we evaluate whether contact information inferred by \nameMethod can be used to improve human pose and shape (\hps) regression; we do so
using the PROX ``quantitative" dataset \cite{Hassan2019prox}. PROX uses an optimization method to fit \smplx bodies to images. It further assumes a-priori known \threeD scenes and uses manually-annotated contact regions on the body to encourage these body vertices to be in contact with the scene if they are sufficiently close, while penalizing body-scene penetration.}

\camready{
Specifically, we replace the manually-annotated contact vertices with the inferred \smplx body-part contact vertices from baseline methods as well as the detailed contact estimated by \nameMethod. 
For a fair comparison, we follow the same experimental setup as HOT \cite{HOT} and evaluate all methods using the Vertex-to-Vertex (V2V) error. For the ``No contact" setup, we turn off all contact constraints in the optimization process. PROX uses the contact regions on the body from the original method~\cite{Hassan2019prox}. HOT uses the body-part vertices from the body-part labels predicted by the HOT detector. We also report V2V errors when using the ground-truth (GT) contact vertices. 
The results in \cref{tab:supmat:hps_downstream} illustrate the value of inferring detailed contact on the body.} 

\camready{
All baselines in \cref{tab:supmat:hps_downstream} use PROX's~\cite{Hassan2019prox} hyperparameters for a fair comparison. 
PROX uses a Geman-McClure robust error function (GMoF) for the contact term (see Eq.4 in \cite{Hassan2019prox}), so that the manually-defined contact areas that lie ``close enough" to the scene are snapped onto it. 
The robust scale term, $\rho_C=5e-02$, is tuned for PROX's naive contact prediction; 
this is relatively conservative as PROX uses no image contact for this prediction. 
Since \nameMethod takes into account the image features, and makes a much more informed contact prediction, 
we we can ``relax'' this robustness term, and trust the output of DECO regressor more.
In \cref{tab:supmat:gmof} we report a sensitivity analysis by varying $\rho_C$ with \nameMethod's contact predictions. 
The results verify that we can trust DECO's contacts more, and that there is a sweet spot for $\rho_C=1.0$. 
This suggests that exploiting inferred contact is a promising direction for improving \hps estimates.}

\begin{table}[t]
\centering
\resizebox{\columnwidth}{!}{
     \begin{tabular}{l | c  c  c  c  c  c}
        \toprule
        ~\textbf{Method}~   & ~No~    & ~PROX~  &  ~HOT~ &  ~\nameMethod~  &  ~GT~     \\
        ~~                  & ~Contact~ & \cite{Hassan2019prox} &   \cite{HOT}  &   ~Contact~ &  ~Contact~  \\
        \midrule 
        V2V $\downarrow$    & 183.3 &	174.0   &   172.3  & \textbf{171.6}  & 163.0 \\
        \bottomrule 
     \end{tabular}
 }
\vspace{-0.5em}
\caption{\hps estimation performance using contact derived from different sources.
}
\label{tab:supmat:hps_downstream}
\end{table}

\begin{table}[t]
\centering
\resizebox{\columnwidth}{!}{
     \begin{tabular}{l | c  c  c  c  c  c  c  c}
        \toprule
        ~\textbf{GMoF $\rho_C$}~   & 1e-03    & 5e-02  &  1e-01 &  1.0  &  2.0  &  3.0   &  5.0 \\
        \midrule 
        V2V $\downarrow$    & 180.07 &	171.6   &  170.0  & {\bf 169.0}  & 176.5  &  179.6  & 183.5 \\
        \bottomrule 
     \end{tabular}
 }
\vspace{-0.5em}
\caption{Sensitivity analysis for the $\rho_C$ value in the Geman-McClure error function (GMoF) of the contact term. 
}
\label{tab:supmat:gmof}
\end{table}

\section{Conclusion}
\label{sec:conclusion}

We focus on detecting 3D human-object contact from a single image taken in the wild; existing methods perform poorly for such images.
To this end, we use crowd-sourcing to collect \nameDataset, a rich dataset of in-the-wild images paired with pseudo ground-truth 3D contacts on the vertex level, as well as labels for the involved objects and body parts. 
Using \nameDataset, we train \nameMethod, a novel model that detects contact on a 3D body from a single color image. 
\nameMethod's novelty lies in cross-attending to both the relevant body parts and scene elements, while it also anchors the inferred 3D contacts to the relevant 2D pixels. 
Experiments show that \nameMethod outperforms existing work by a good margin, and generalizes reasonably well in the wild.
To enable further research, 
\camready{we release our data, models and code.}

\zheading{Future work}
\nameMethod currently reasons about contact between a single person, the scene, and multiple objects.
Our labelling tool and \nameMethod could be extended to fine-grained human-human, human-animal and self-contact.
Another promising, but challenging, direction would be to leverage captions in existing datasets, or methods that infer captions for unlabeled images, via large language models (LLM).

\label{sec:acknowledgements}
\camready{
{%
\qheading{Acknowledgements} 
We sincerely thank Alpar Cseke for his contributions to DAMON data collection and PHOSA evaluations, Sai K. Dwivedi for facilitating PROX downstream experiments, Xianghui Xie for help with CHORE evaluations, Lea M\"uller for her help in initiating the contact annotation tool, Chun-Hao P. Huang for RICH discussions and Yixin Chen for details about the HOT paper. We are grateful to Mengqin Xue and Zhenyu Lou for their collaboration in BEHAVE evaluations 
and Tsvetelina Alexiadis for valuable data collection guidance. Their invaluable contributions enriched this research significantly. This work was funded by the International Max Planck Research School for Intelligent Systems (IMPRS-IS).
}
\noindent \small \textbf{{Disclosure}:} \href{https://files.is.tue.mpg.de/black/CoI_ICCV_2023.txt}{https://files.is.tue.mpg.de/black/CoI\_ICCV\_2023.txt} 
}

\appendix
{\noindent\LARGE\textbf{Supplementary Material}}
\newline
\renewcommand{\thefigure}{S.\arabic{figure}}
\renewcommand{\thetable}{S.\arabic{table}}
\renewcommand{\theequation}{S.\arabic{equation}}
\setcounter{figure}{0}
\setcounter{table}{0}
\setcounter{equation}{0}

\section{\nameDataset Data Collection and Quality}

We select images for annotation from the HOT~\cite{HOT} curated subset of V-COCO~\cite{gupta2015vcoco} and HAKE~\cite{li2020hake} by filtering out images containing multiple people or images with a single person but fewer than $10$ visible keypoints. For keypoint estimation, we use the transformer-based SOTA \twoD keypoint estimator ViTPose~\cite{xu2022vitpose}.

\camready{We take several steps to limit ambiguity in the contact annotation task. Here, we focus on \emph{scene-} and \emph{human-supported} contact.
The requirement for support resolves ambiguous cases, \eg humans close to scene objects but not in contact. We use the object labels in V-COCO and HAKE to filter out images containing unsupported human-human and human-animal contact. V-COCO and HAKE also contain action labels that we leverage to filter out ambiguous indirect contact which does not involve physical touch, such as direct, greet, herd, hose, point, teach,~\etc. The training video (in \supmat) advises workers to orient the \threeD mesh and to visualize themselves in the same posture as the person in the image. 
This 
helps 
infer contact while avoiding left-right ambiguity. 
Our 
Fleiss' Kappa score indicates significant agreement between annotators (see \cref{quality_control_and_eval}), suggesting that our protocol effectively minimizes 
task ambiguities.}

To facilitate crowd-sourced \threeD contact annotation using Amazon Mechanical Turk (AMT), we build a new annotation tool which we describe in detail in the following section. 
Please see the \textbf{Supplemental Video}.

\subsection{Dense Contact Annotation Tool}

\begin{figure*}[ht]
\centering
\includegraphics[width=\linewidth]{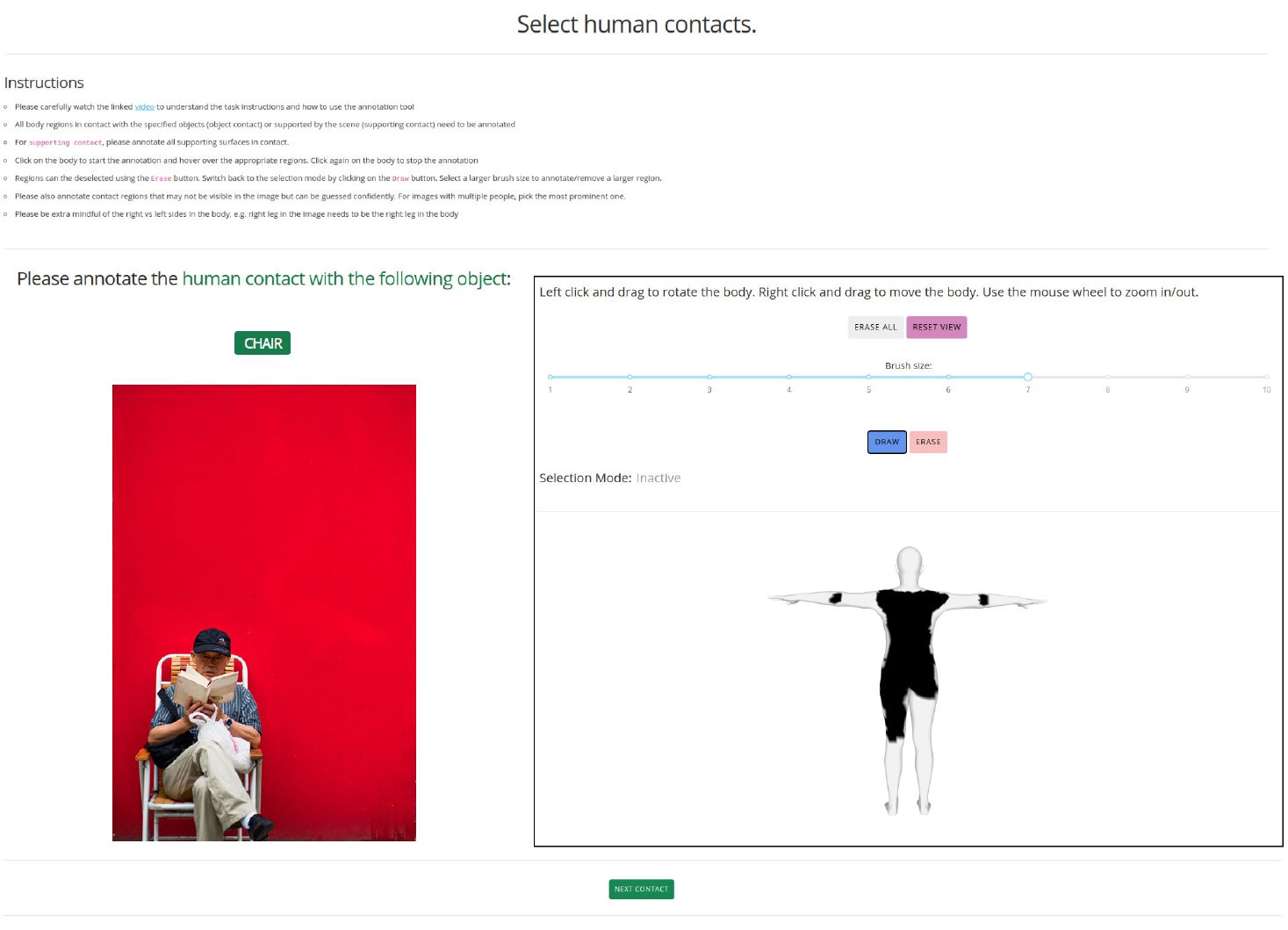}
\caption{AMT interface design for our annotation tool. We show a an example of an annotator collecting \textit{human-supported} contact for the object label ``Book". The application cycles through all available object labels in the image and the \textit{scene-supported} contact. Please refer to the \textbf{Supplemental Video} for a detailed description of the tool. \faSearch~\textbf{Zoom in}}
\label{figure:supmat:annotation_interface}
\end{figure*}

We built a dense contact annotation tool to collect annotations from the \nameDataset dataset images.
The code for the tool is written using \href{https://github.com/plotly/dash}{Dash},
a popular Python framework for building web applications. 
This application is deployed inside a \href{https://www.docker.com/}{Docker} container 
under an \href{https://uwsgi-docs.readthedocs.io/en/latest/}{uWSGI} application server, eventually served by a \href{https://www.nginx.com/}{NGINX}
web server acting as a reverse proxy.
The annotation tool is accessible under a public URL used to create
the Human Intelligent Tasks on AMT. 

\textbf{Interface and use.}
As seen in Fig.~\ref{figure:supmat:annotation_interface}, 
the application is made of four parts. 
The top part contains a title and general instructions about how to use the annotation tool.
The left part is made of the image and a label describing which contact 
should be annotated (object or supporting contact).
The right part contains the mesh to be annotated by hovering over it. 
The mesh can be translated, rotated, and zoomed-in/out.
A slider allows the user to select the size of the brush, and buttons are available
for switching modes (draw/erase), erasing the full selection, 
and resetting the camera.
Finally, a confirmation button is located at the bottom of the window
to submit an annotation to the server.
The user must provide one annotation for several human-object contacts
and for the supporting contact.
Once the last annotation has been submitted, 
a dialog box appears to ask for optional feedback about the annotation task for the current image. This helps workers report ambiguous contact scenarios.

\textbf{Callbacks.}
Dash applications work with \textit{callbacks}. Callbacks are functions that are 
fired when an input component is updated (e.g., a button is clicked) 
and that update output components. \textit{Regular} callbacks are executed on the 
server-side: they are simpler to implement, but slower to execute. On the other hand, 
\textit{client-side} callbacks are faster but require a more complex implementation.
The user will spend most of their time annotating the high-resolution mesh.
It should therefore be smooth and fast.
As such, we implemented this logic in JavaScript as a client-side callback. 
Other callbacks, for instance when the camera is reset or the brush size is updated, 
rarely happen and do not require a fast response. Therefore, they have been 
implemented as server-side callbacks. During their execution, a spinner appears
to let the worker know that the application is updating.

\textbf{Caching.}
When a vertex is annotated, vertices belonging to a neighboring region are also annotated.
The extent of this neighboring region is correlated with the brush size that the
user chooses. When we start the application, we compute, for each vertex and for each brush size, 
all of its neighboring vertices. As the mesh is static, this has to be done only once.
Therefore, we cache this result and use it for all annotations.

\textbf{Video.}
Please watch the  \textbf{Supplementary Video} for an in-depth tour of our tool, its features and the annotation protocol. Note that this is the same video we showed AMT workers for training purposes during qualification.

\subsection{\nameDataset Additional Statistics}

\Cref{figure:supmat:damon_object_stats_full} shows the full version of Fig.~3 in the main paper.  The \nameDataset dataset is long-tailed and it covers contact scenarios with a wide variety of objects and scenes. Please refer to the sunburst plot in \cref{figure:supmat:damon_object_stats_full} for a full breakdown. 

\begin{figure}[h]%
\vspace{-0.6 em}
\hspace{-1,5 em}\includegraphics[trim=80 10 170 240,clip,width=1.09 \linewidth]{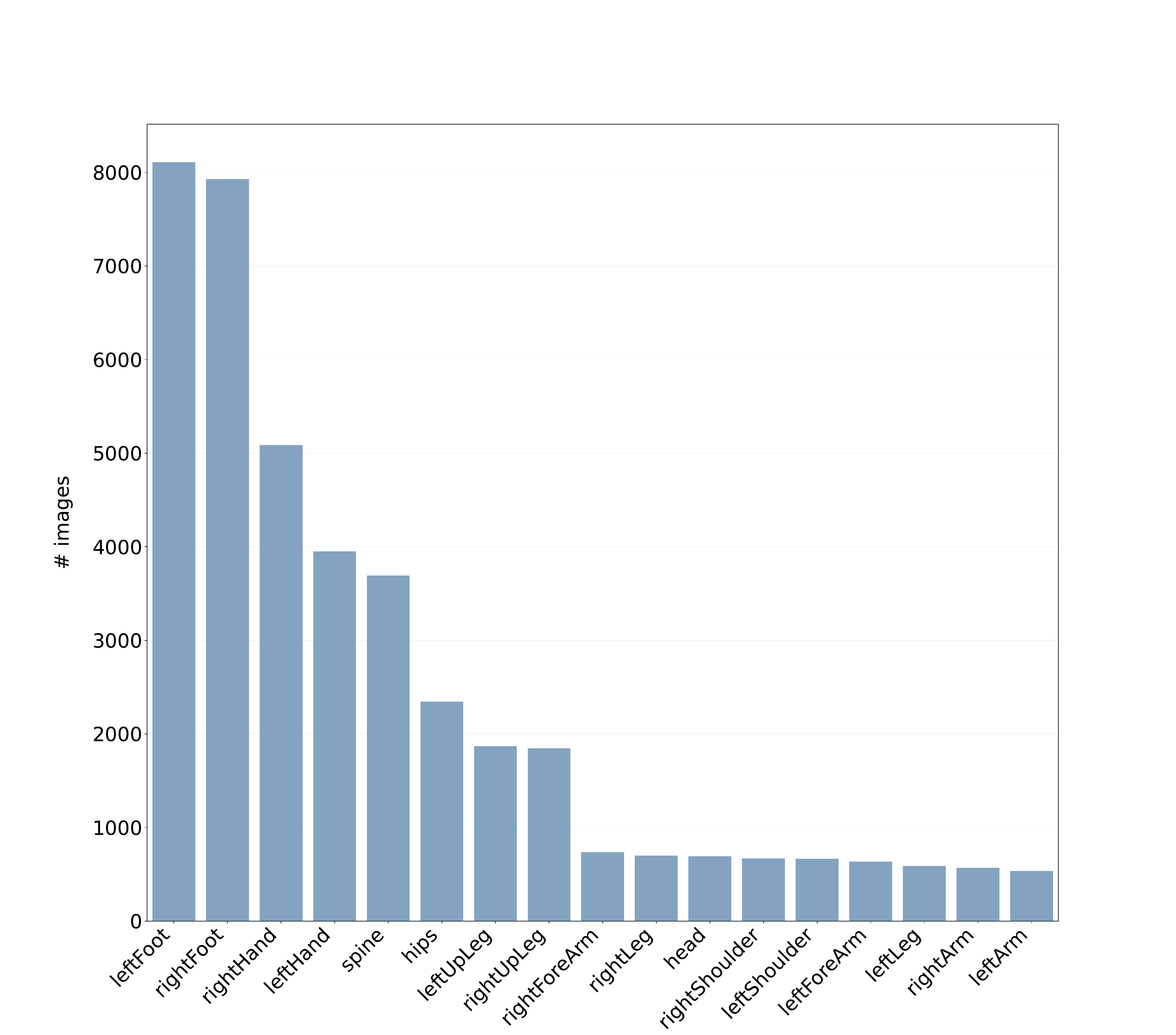}
\vspace{-2.3 em}
\caption{
            \nameDataset dataset statistics. %
            The number of images ($y$-axis) for which each body part ($x$-axis) has at least 10 vertices in contact.
            For visualization purposes here we combine the fingers into the hands category, the toes into the feet category, and several spine parts into a single spine category (SMPL has 24 parts but here we show 17 bars). 
            \faSearch~\textbf{Zoom in}. %
}
\label{figure:damon:statsBodyPart}
\end{figure}

\begin{figure*}[ht]
\centering
\vspace{-0.5 em}%
\includegraphics[width=1.0\linewidth]{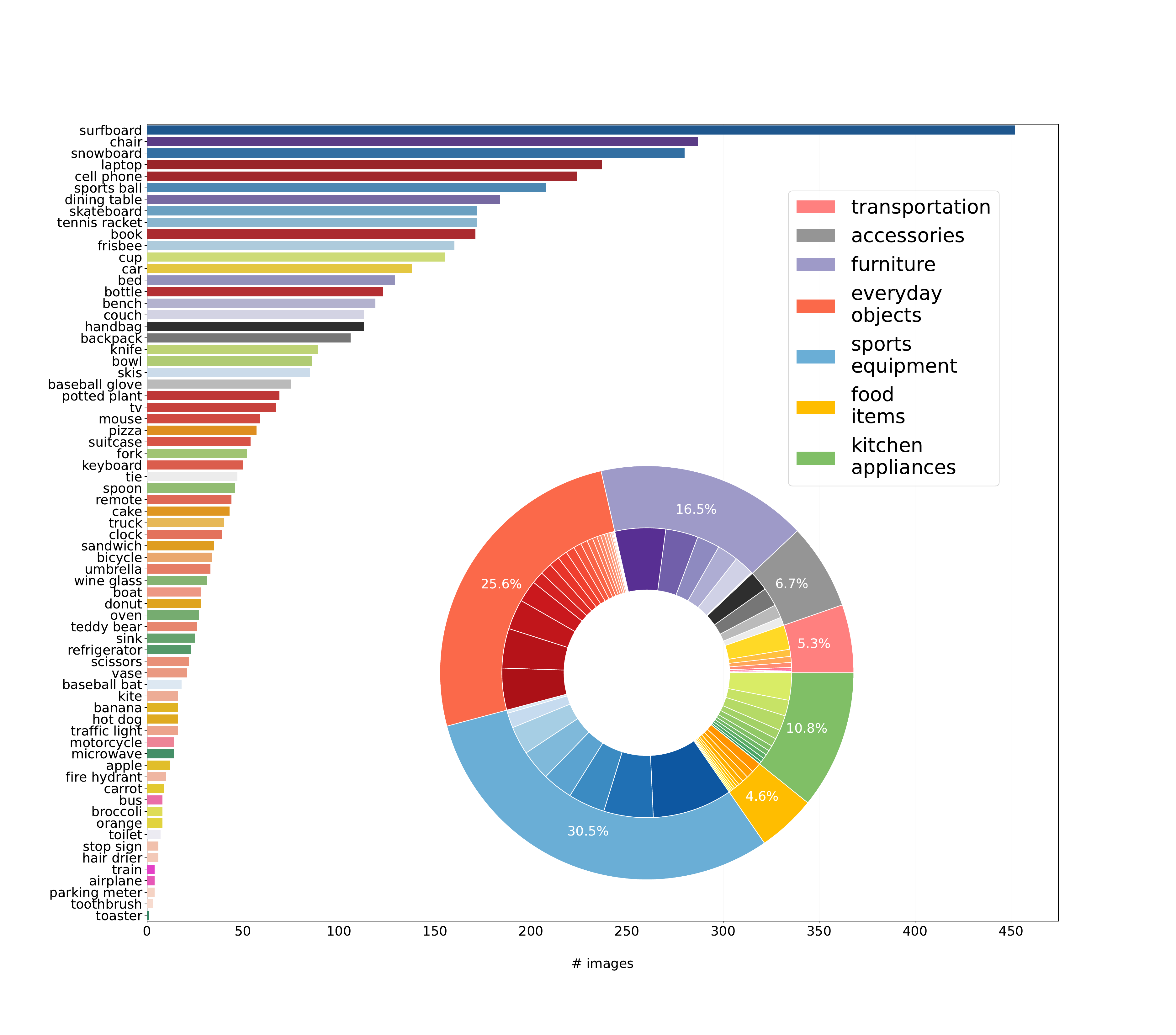}
\vspace{-2.0 em}
\caption{\emph{Full version} plot for \nameDataset dataset statistics (Fig. 3 in Main). %
            \textbf{Histogram}: 
            contact object labels ($y$-axis) and the number of images in which they are present ($x$-axis). 
            \textbf{Pie chart}: 
            object labels are grouped into 7 main categories; 
            inner colors correspond to the colors in the histogram.
            \faSearch~\textbf{Zoom in}.} %
\label{figure:supmat:damon_object_stats_full}
\vspace{-1.0 em}
\end{figure*}

\camready{\Cref{figure:damon:statsBodyPart} 
shows the number of images per object label.
We see that contact with feet, hands, and the bigger body parts (torso, hips, upper arms) prevails; this makes sense as humans interact with objects mostly with these (\eg, for walking, grasping, sitting, lying down). 
However, interactions are highly varied, thus, the distribution is long-tailed and includes all body parts.}

\camready{Workers take on average 3.48 min/image and we pay \$0.5/image. The total cost is \$3313.20 with AMT fees. 
The \nameDataset contact annotations are not prohibitively expensive given that it provides
a stepping stone for future research.}

\begin{figure*}%
\centering
\includegraphics[trim=000mm 000mm 000mm 000mm, clip=true, width=1.0\linewidth]{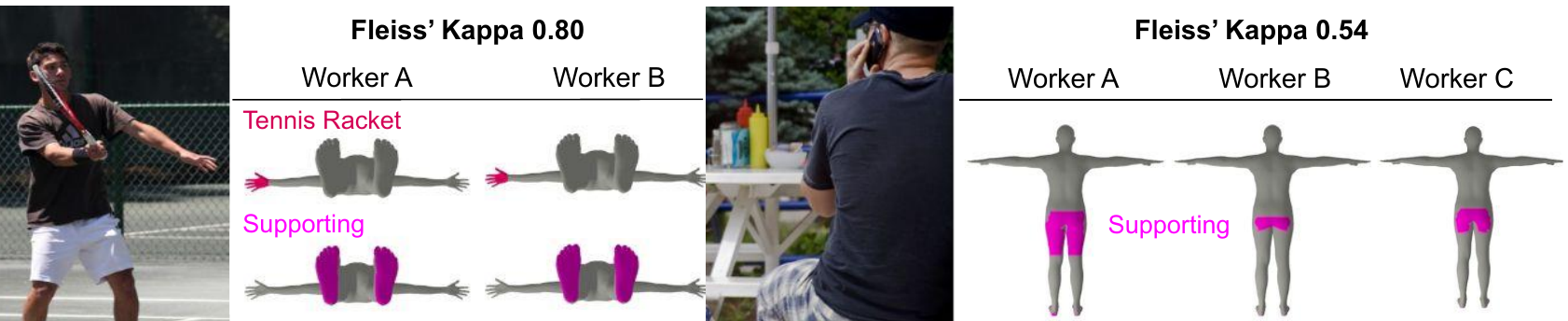}
\caption{Annotator agreement indicated by $\kappa$ \faSearch~~\textbf{Zoom in}.}
\label{figure:supmat:fleiss-kappa-agreement}
\end{figure*}

\subsection{Quality Control and Evaluation}
\label{quality_control_and_eval}

We adopt two strategies to \camready{ensure quality and} avoid noisy annotations in the \nameDataset dataset. First, we conduct qualification tasks to shortlist high-quality annotator candidates. This qualification task has two parts: (i) watching a detailed tutorial video (see \textbf{Supplementary Video}) explaining the task and annotator tool step-by-step by showing three example annotations with varying degrees of contact complexity, (ii) annotating 10 sample images for contact annotations. For the sample images, we had a set of \camready{author-annotated} \emph{pseudo-ground-truth} (pseudo GT) labels. The responses of candidates were evaluated using Intersection-over-Union (IoU) with the pseudo-GT labels. Workers who responded satisfactorily were allowed to annotate the \nameDataset dataset images. We qualified 14 out of 100 participants after the qualification round. The second strategy involved hiring Master's students \camready{as meta-annotators} to visually inspect the quality of contact annotations. Annotations that were flagged as incorrect \camready{or low-quality} were sent for re-annotation with specific feedback to the annotators on how to avoid mistakes. 

We assess the quality of the \nameDataset dataset by measuring the \emph{label accuracy} and the \emph{level of annotator's agreement}. 

\begin{figure*}[ht]
\centering
\vspace{-0.5 em}%
\includegraphics[width=1.0\linewidth]{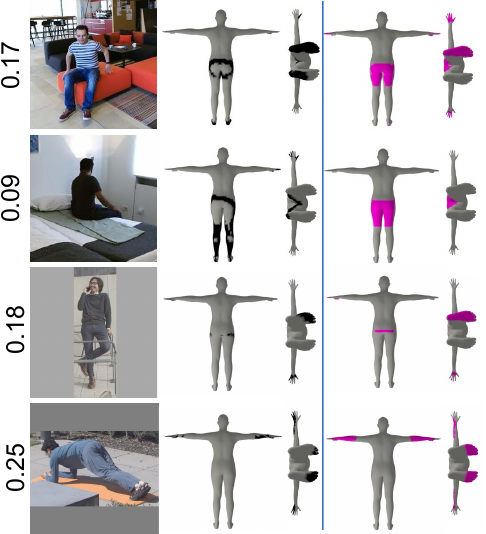}
\vspace{-2.0 em}
\caption{\nameDataset annotations (in \textcolor{magenta}{magenta}) earning the lowest IOU scores compared to GT contact in PROX and RICH (in \textbf{black}). IOU scores are reported to the left of RGB images for each row. Scanned datasets (\eg PROX/RICH) infer contact by thresholding the SDF between body and scene, which can be sub-optimal due to soft-tissue deformation of the body (see text).}
\label{figure:supmat:bad_gt_iou}
\vspace{-1.0 em}
\end{figure*}

We evaluate label accuracy by manually selecting 100 images with contact labels from the RICH~\cite{RICH} and PROX~\cite{Hassan2019prox} datasets. Note that the pseudo-ground-truth contact labels in these datasets are obtained by thresholding the Signed Distance Field (SDF) between the reconstructed human mesh and the \threeD scene. We evaluate annotations from qualified workers on these images and compute IoU \wrt the pseudo-ground-truth contact labels. With this, we obtain an IOU score of 0.512 on RICH, 0.263 on PROX, and a mean IOU (mIOU) score of 0.450. 

\Cref{figure:supmat:bad_gt_iou} visualizes the \nameDataset annotation earning the lowest IoU scores. Scanned datasets that rely on thresholding SDF values for estimating contact labels fail to take into account the soft-tissue deformation of the human body when it interacts with rigid objects. The vertices in the ``soft'' body parts such as buttocks, thighs, etc interpenetrate far enough from the scan surface to overshoot the heuristic threshold, leading to noisy GT annotation and a ``ring'' like contact profile. \nameDataset is annotated by human annotators and therefore does not suffer from this issue. 
This produces a mismatch between these two types of ground truth. 
Note that \nameDataset ground truth is closer to reality.

\camready{We also compare annotations on a randomly-selected set of 10 images from all the qualified workers against author-annotated labels, resulting in mIOU = $0.510$.} 

To determine the agreement between annotators, qualified workers annotate the same set of 10 images and we report the \emph{Fleiss' Kappa ($\kappa$)} metric. Fleiss' Kappa is a statistical measure used to evaluate the agreement level among a fixed number of annotators when assigning categorical labels to data.  It considers the possibility of chance agreement and provides a standardized measure of inter-rater reliability that ranges from 0 (no agreement) to 1 (perfect agreement). In this study, we obtain a Fleiss' Kappa $\kappa=0.656$ which is considered ``substantial agreement" between workers~\cite{gwet2014handbook}. Note, $\kappa$ of 1 means ``perfect agreement'', 0 means ``chance agreement'' and -1 means ``perfect disagreement''. 
\camready{To build intuition on the significance of $\kappa$, \cref{figure:supmat:fleiss-kappa-agreement} shows example annotations %
with low and high $\kappa$ scores.}

\section{\nameMethod Experiments}

\subsection{Implementation Details}

For training \nameMethod, we resize input images, the scene segmentation mask and the part segmentation mask such that $\mathbf{I} \in \mathbb{R}^{3\times 256\times 256}$, $\boldsymbol{X}_s \in \mathbb{R}^{133\times 256\times 256}$ and $\boldsymbol{X}_p \in \mathbb{R}^{25\times 256\times 256}$. $\mathbf{F}_s$ and $\mathbf{F}_s$ are of size $\mathbb{R}^{480\times64\times64}$S. We determine the loss weights in Eqn. 4 empirically and set it to $w_c=10.0$, $w_{pal}=0.05$, $w_s=1.0$ and $w_p=1.0$. We use the Adam optimizer with a learning rate of $5 \times 10^{-5}$ and batch size of 4, and training takes 12 epochs ($\sim 31$ hours) on an Nvidia Tesla V100 GPU.

For evaluation on RICH-test in Tab. 1 in main, we sub-sample every 10th frame from the released test set.

\camready{The base model without context branches has 90.19M parameters. 
Adding context branches (${\mathcal{L}_s^{2D}}$ and ${\mathcal{L}_p^{2D}}$)
adds another 
853K parameters. %
This improves the geodesic error by $\sim$24\% (see Tab.~\textcolor{red}{1} in main), at the cost of $\sim$1\% increase in complexity. 
We will release both models, with and without context branches.}

\subsection{Additional Qualitative results}

\definecolor{ContactGreen}{rgb}{0.1, 0.8, 0.1}
\begin{figure*}[ht]
\centering %
\includegraphics[width=0.85\linewidth]{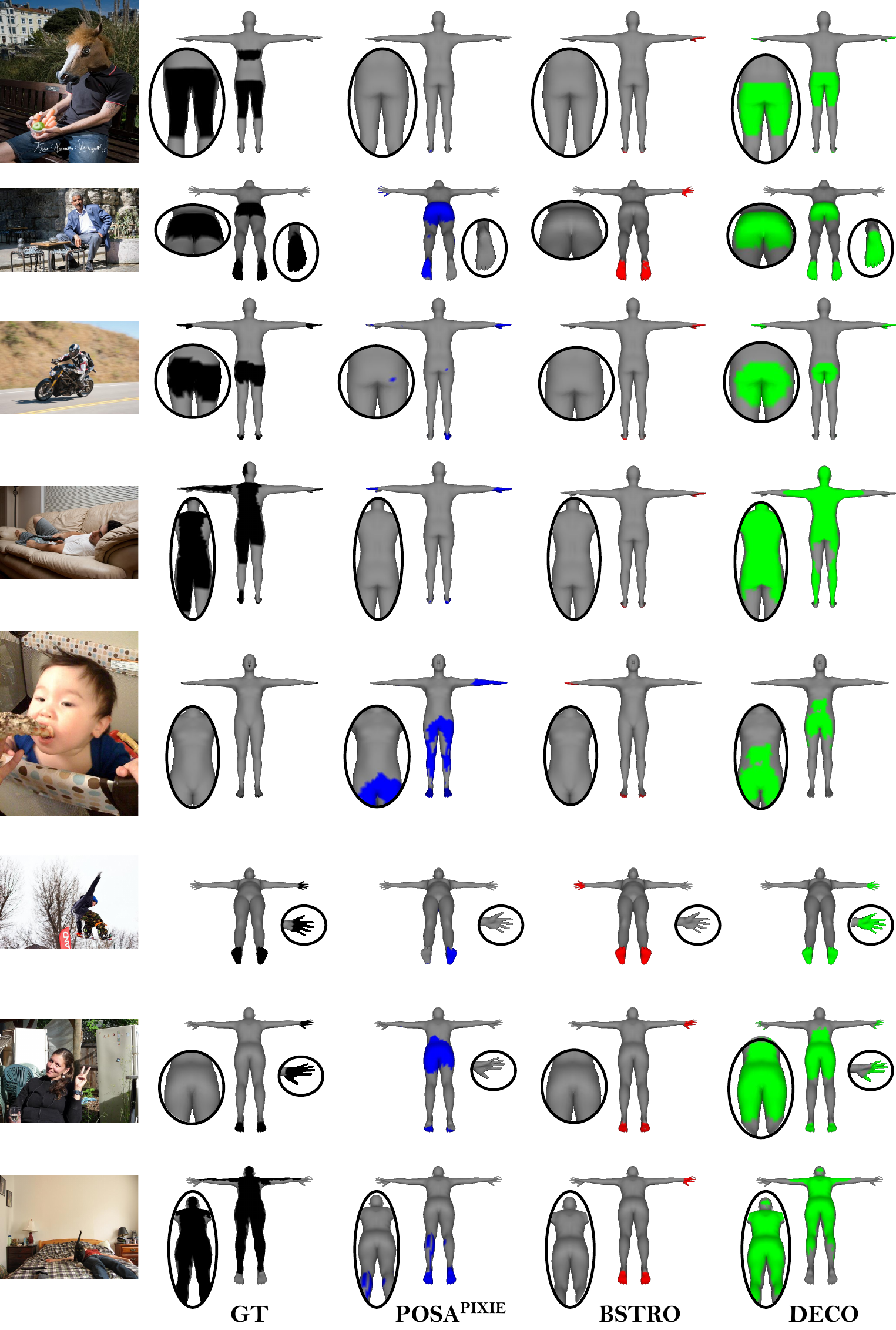}
\caption{Additional qualitative evaluation of 
            \nameMethod             (\textcolor{ContactGreen}{green}), 
            BSTRO                   (\textcolor{red}{red}) and 
            POSA$^{\text{PIXIE}}$   (\textcolor{blue}{blue}), 
            alongside Ground Truth  (\textbf{black}) on images from the \nameDataset dataset. %
}
\label{figure:supmat:deco_qual_comparison_hot}
\end{figure*}

\definecolor{ContactGreen}{rgb}{0.1, 0.8, 0.1}
\begin{figure*}[ht]
\centering %
\includegraphics[width=\linewidth]{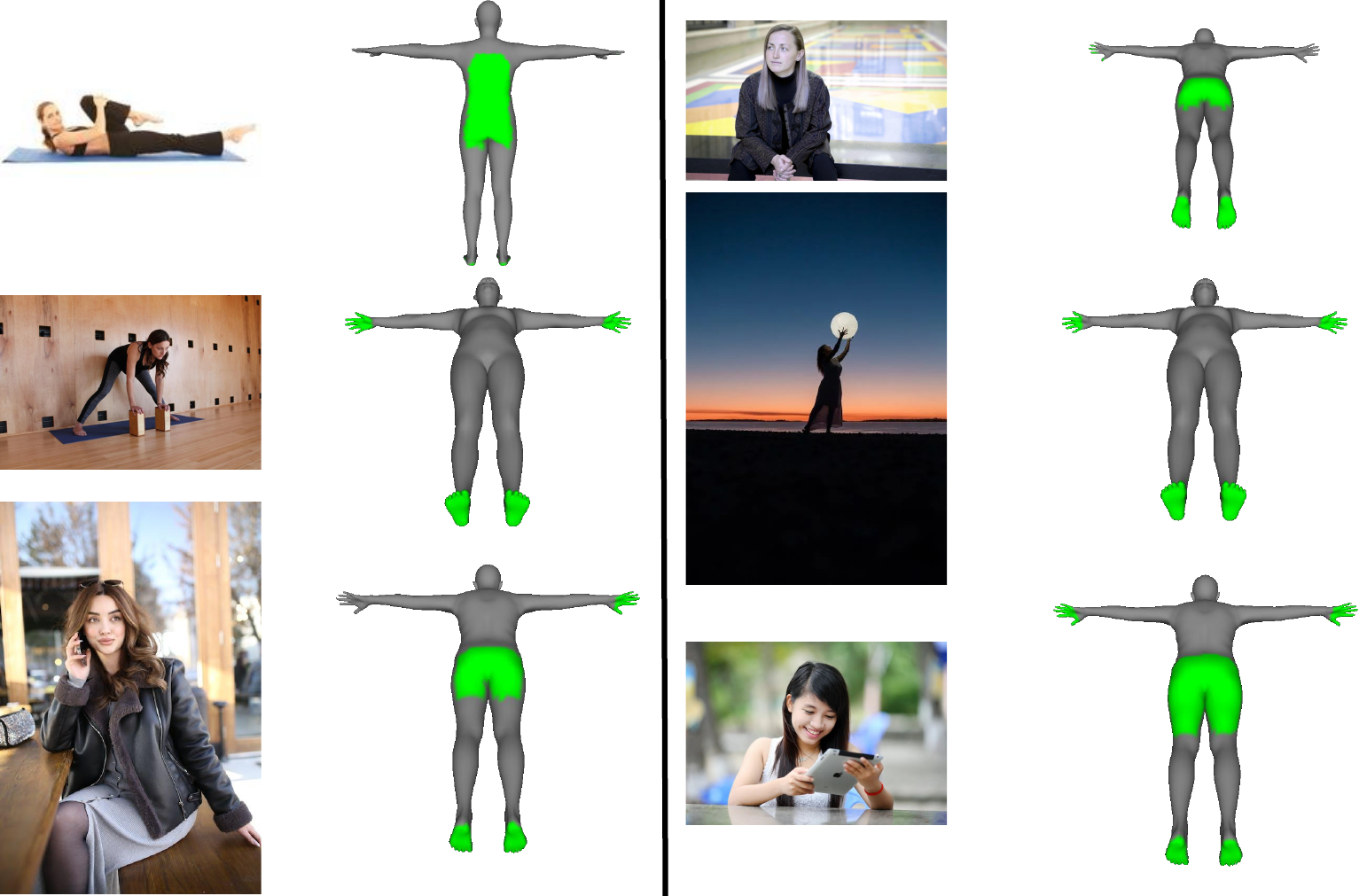}
\caption{\nameMethod predictions (in \textcolor{ContactGreen}{green}) on Internet images, not seen during training.}
\label{figure:supmat:deco_qual_comparison_internet}
\end{figure*}

\Cref{figure:supmat:deco_qual_comparison_hot} shows \nameMethod estimated contact and comparison with baseline methods from the test subset of \nameDataset. \Cref{figure:supmat:deco_qual_comparison_internet} shows \nameMethod contacts on some randomly sampled images from the internet.

\clearpage

{\small
\balance
\bibliographystyle{config/iccv2023/ieee_fullname}
\bibliography{config/BIB}
}

\end{document}